\def\year{2022}\relax
\documentclass[letterpaper]{article} 
\usepackage{aaai22}  
\usepackage{times}  
\usepackage{helvet}  
\usepackage{courier}  
\usepackage[hyphens]{url}  
\usepackage{graphicx} 
\def\UrlFont{\rm}  
\usepackage{natbib}  
\usepackage{caption} 
\DeclareCaptionStyle{ruled}{labelfont=normalfont,labelsep=colon,strut=off} 
\usepackage{amsmath,amsfonts,bm}

\def\eqref#1{equation~\ref{#1}}

\def\1{\bm{1}}

\DeclareMathAlphabet{\mathsfit}{\encodingdefault}{\sfdefault}{m}{sl}
\SetMathAlphabet{\mathsfit}{bold}{\encodingdefault}{\sfdefault}{bx}{n}

\newcommand{\E}{\mathbb{E}}

\newcommand{\Var}{\mathrm{Var}}

\usepackage[utf8]{inputenc} 
\usepackage[T1]{fontenc}    
\usepackage{url}            
\usepackage{booktabs}       
\usepackage{amsfonts}       
\usepackage{nicefrac}       
\usepackage{microtype}      

\usepackage{amsmath}
\usepackage{amssymb}
\usepackage{bbm}
\usepackage[capitalise]{cleveref}
\usepackage[bottom]{footmisc} 
\crefformat{section}{\S#2#1#3}
\crefmultiformat{section}{\S#2#1#3}{ and~\S#2#1#3}{, \S#2#1#3}{, and~\S#2#1#3}
\usepackage[raise]{engord}
\usepackage[acronym]{glossaries}
\usepackage{multirow}
\usepackage{rotating}
\usepackage{subcaption}
\usepackage{threeparttable}
\usepackage[separate-uncertainty=true,tight-spacing=true]{siunitx}
\usepackage{textcomp}
\usepackage{adjustbox}
\usepackage{placeins} 

\renewcommand*{\thefootnote}{\fnsymbol{footnote}}

\DeclareMathOperator{\mean}{mean}

\DeclareMathOperator{\FanIn}{{\textit{fan-in}}}
\DeclareMathOperator{\FanOut}{{\textit{fan-out}}}

\glsdisablehyper
\newacronym{dnn}{DNN}{Deep Neural Network}
\newacronym{cnn}{CNN}{Convolutional Neural Network}
\newacronym{nn}{NN}{Neural Network}
\newacronym{lt}{LT}{Lottery Ticket}
\newacronym{lth}{LTH}{Lottery Ticket Hypothesis}
\newacronym{wt}{WT}{Winning Ticket}
\newacronym{jsd}{JSD}{Jensen–Shannon Divergence}
\newacronym{kl}{KL}{Kullback–Leibler Divergence}
\newacronym{ml}{ML}{Machine Learning}
\newacronym{mds}{MDS}{Multi-dimensional Scaling}
\newacronym{set}{SET}{Sparse Evolutionary Training}
\newacronym{rigl}{RigL}{Rigged Lottery}
\newacronym{grasp}{GRaSP}{Gradient Signal Preservation}
\newacronym{snip}{SNIP}{Single-shot Network Pruning}
\newacronym{dst}{DST}{Dynamic Sparse Training}
\newacronym{batchnorm}{BatchNorm}{Batch Normalization}

\newcommand{\imgnet}{ImageNet-2012}

\title{Gradient Flow in Sparse Neural Networks and How Lottery Tickets Win}
\author {
    Utku Evci\equalcontrib\textsuperscript{\rm 1},
    Yani Ioannou\equalcontrib\textsuperscript{\rm 2},
    Cem Keskin \textsuperscript{\rm 3},
    Yann Dauphin \textsuperscript{\rm 1}
}
\affiliations {
    \textsuperscript{\rm 1} Google,
    \textsuperscript{\rm 2} University of Calgary,
    \textsuperscript{\rm 3} Facebook\\
    evcu@google.com, yani.ioannou@ucalgary.ca, cem.keskin@fb.com, ynd@google.com
}

\begin{document}

\maketitle

\begin{abstract}
Sparse \glspl{nn} can match the generalization of dense \glspl{nn} using a fraction of the compute/storage for inference, and have the potential to enable efficient training. However, naively training unstructured sparse \glspl{nn} from random initialization results in significantly worse generalization, with the notable exceptions of \glspl{lt} and \gls{dst}. Through our analysis of gradient flow during training we attempt to answer: (1) why training unstructured sparse networks from random initialization performs poorly and;  (2) what makes \glspl{lt} and \gls{dst} the exceptions? We show that sparse \glspl{nn} have poor gradient flow \emph{at initialization} and demonstrate the importance of using sparsity-aware initialization.
Furthermore, we find that \Gls{dst} methods significantly improve gradient flow \emph{during training} over traditional sparse training methods. Finally, we show that \Glspl{lt} do not improve gradient flow, rather their success lies in re-learning the pruning solution they are derived from --- however, this comes at the cost of learning novel solutions.
\end{abstract}
\glsresetall

\section{Introduction}
\label{sec:intro}
\Glspl{dnn} are the state-of-the-art method for solving problems in computer vision, speech recognition, and many other fields. While early research in deep learning focused on application to new problems, or pushing state-of-the-art performance with ever larger/more computationally expensive models, a broader focus has emerged towards their efficient real-world application. One such focus is on the observation that only a sparse subset of this dense connectivity is required for inference, as apparent in the success of \emph{pruning}. 

Pruning has a long history in \gls{nn} literature~\citep{mozer1989skeletonization,han2015learning}, and remains the most popular approach for finding sparse \glspl{nn}. Sparse \glspl{nn} found by pruning algorithms~\citep{han2015learning,gupta2018,Molchanov2017,louizos2017bayesian} (i.e.\ \emph{pruning solutions}) can match dense \gls{nn} generalization with much better efficiency at \emph{inference} time.
However, naively \emph{training} an (unstructured) sparse \gls{nn} from a random initialization (i.e.\ \emph{from scratch}), typically leads to significantly worse generalization.

Two methods in particular have shown some success at addressing this problem --- \glspl{lt} and \gls{dst}. However, we don't know how to find \Acrfullpl{lt} efficiently; while RigL~\citep{evci2019b}, a recent \gls{dst} method, requires 5$\times$ the training steps to match dense \gls{nn} generalization. Only in understanding how these methods overcome the difficulty of sparse training can we improve upon them.

A significant breakthrough in training \glspl{dnn} --- addressing vanishing and exploding gradients --- arose from understanding gradient flow both at initialization, and during training.
In this work we investigate the role of gradient flow in the difficulty of training unstructured sparse \glspl{nn} from random initializations and from \gls{lt} initializations. Our experimental investigation results in the following insights:
\begin{enumerate}
\item \textbf{Sparse \glspl{nn} have poor gradient flow at initialization.} In \cref{sec:methods_init}, \cref{sec:exp_init} we show that the predominant method for initializing sparse \glspl{nn} is incorrect in not considering heterogeneous connectivity. We believe we are the first to show that sparsity-aware initialization methods improve gradient flow and training. 
\item \textbf{Sparse \glspl{nn} have poor gradient flow during training.} In \cref{sec:methods_dst}, \cref{sec:exp_dst}, we observe that even in sparse \gls{nn} architectures less sensitive to incorrect initialization, the gradient flow \emph{during training} is poor. We show that \gls{dst} methods achieving the best generalization have improved gradient flow, especially in early training.
\item \textbf{\acrlongpl{lt} don't improve upon (1) or (2), instead they re-learn the pruning solution.} In \cref{sec:methods_lottery}, \cref{sec:exp_lottery} we show that a \gls{lt} initialization resides within the same basin of attraction as the original pruning solution it is derived from, and a \gls{lt} solution is highly similar to the pruning solution in function space.
\end{enumerate}

\section{Related Work}
\label{related}
\paragraph{Pruning} Pruning is used commonly in \gls{nn} literature to obtain sparse networks~\citep{mozer1989skeletonization,han2015learning,Kusupati2020str}. While the majority of pruning algorithms focus on pruning \emph{after} training, a subset focuses on pruning \glspl{nn} \emph{before} training \citep{snip,grasp,tanaka2020pruning}. \Gls{grasp}~\citep{grasp} is particularly relevant to our study, since their pruning criteria aims to preserve gradient flow, and they observe a positive correlation between gradient flow \textit{at initialization} and final generalization. However, the recent work of \citet{Frankle2020pruningatinit} suggests that the reported gains are due to sparsity distributions discovered rather than the particular sub-network identified. Another limitation of these algorithms is that they don't scale to large-scale tasks like ResNet-50 training on \imgnet.

\paragraph{Lottery Tickets}
\Citet{Frankle2018lottery} showed the existence of sparse sub-networks at initialization ---  known as \Acrlongpl{lt} ---  which can be trained to match the generalization of the corresponding dense \gls{dnn}.
The initial work of \citet{Frankle2018lottery} inspired much follow-up work. 
\Citet{Liu2018,gale2019state} observed that the initial formulation was not applicable to larger networks with higher learning rates. \citet{Frankle2019stabilizing,Frankle2019linearmode} proposed \textit{late rewinding} as a solution. 
\Citet{Morcos2019oneticket, Sabatelli2020transferability} showed that \glspl{lt} trained on large datasets transfer to smaller ones, but not \emph{vice versa}. 
\Citet{Zhou2019deconstructing,Frankle2020early,rananujan2019} focused on further understanding \glspl{lt}, and finding sparse sub-networks at initialization.
However, it is an open question whether finding such networks at initialization could be done more efficiently than with existing pruning algorithms.

\paragraph{Dynamic Sparse Training} Most training algorithms work on pre-determined architectures and optimize parameters using fixed learning schedules. \Acrfull{dst}, on the other hand, aims to optimize the sparse \gls{nn} connectivity jointly with model parameters. \citet{Mocanu2018,Mostafa2019} propose replacing low magnitude parameters with random connections and report improved generalization. \Citet{dettmers2019} proposed using momentum values, whereas \citet{evci2019b} used gradient estimates directly to guide the selection of new connections, reporting results that are on par with pruning algorithms, and has been applied to vision transformers \citep{Chen2021ChasingSI}, language models \citep{dietrich2021}, reinforcement learning \citep{sokar2021}, training recurrent neural networks \citep{Liu2021EfficientAE} and fast ensembles \citep{freetickets2021}. In \cref{sec:exp_dst} we study these algorithms and try to understand the role of gradient flow in their success.
\paragraph{Random Initialization of Sparse \gls{nn}} In training sparse \gls{nn} from scratch, the vast majority of pre-existing work on training sparse \gls{nn} has used the common initialization methods~\citep{Glorot2010understanding,He2015delvingdeep} derived for \emph{dense} \glspl{nn}, with only a few notable exceptions. \Citet{Liu2018,rananujan2019,gale2019state} scaled the variance (fan-in/fan-out) of a sparse \gls{nn} layer according to the layer's sparsity, effectively using the standard initialization for a small dense layer with an equivalent number of weights as in the sparse model. \citet{signalprop}  measures the singular values of the input gradients and proposes to use a dense orthogonal initialization to ensure dynamical isometry and improve one-shot pruning performance before training. Similar to our work, \citet{tessera2021, singh2021} compares dense and sparse networks at initialization and during training using gradient flow, whereas \citet{Golubeva2021AreWN} study the distance to the infinite-width kernel. 

\section{Analyzing Gradient Flow in Sparse NNs}\label{sec:methods}
A significant breakthrough in training very deep \glspl{nn} arose in addressing the \emph{vanishing}/\emph{exploding gradient} problem, both at initialization, and during training. This problem was understood by analyzing the signal propagation within a \gls{dnn}, and addressed in improved initialization methods~\citep{Glorot2010understanding,He2015delvingdeep,Lechao2018dynamical} alongside normalization methods, such as \gls{batchnorm}~\citep{Ioffe2015}. In our work, similar to \citet{grasp}, we study these problems using gradient flow, $\nabla L(\theta)^T\nabla L(\theta)$ which is the first order approximation\footnote{We omit learning rate for simplicity and ensure different methods have same learning rate schedules when compared.} of the decrease in the loss expected after a gradient step. We observe poor gradient flow for the predominant sparse \gls{nn} initialization strategy and propose a sparsity-aware generalization in \cref{sec:methods_init}. Then in \cref{sec:methods_dst} and \cref{sec:methods_lottery} we summarize our analysis of \gls{dst} methods and the \gls{lt} hypothesis respectively.
\begin{figure*}[tp]
\centering
\begin{subfigure}{.28\linewidth}
\includegraphics[width=\textwidth]{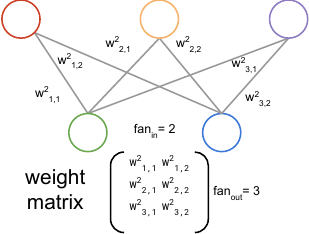}
\caption{Dense Layer}\label{fig:initialization_dense}
\end{subfigure}%
~
\begin{subfigure}{.28\linewidth}
\includegraphics[width=\textwidth]{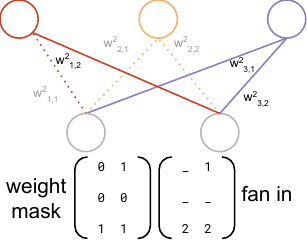}
\caption{Sparse Layer}\label{fig:initialization_sparse}
\end{subfigure}%
~
\begin{subfigure}{.3\linewidth}
\includegraphics[width=\textwidth]{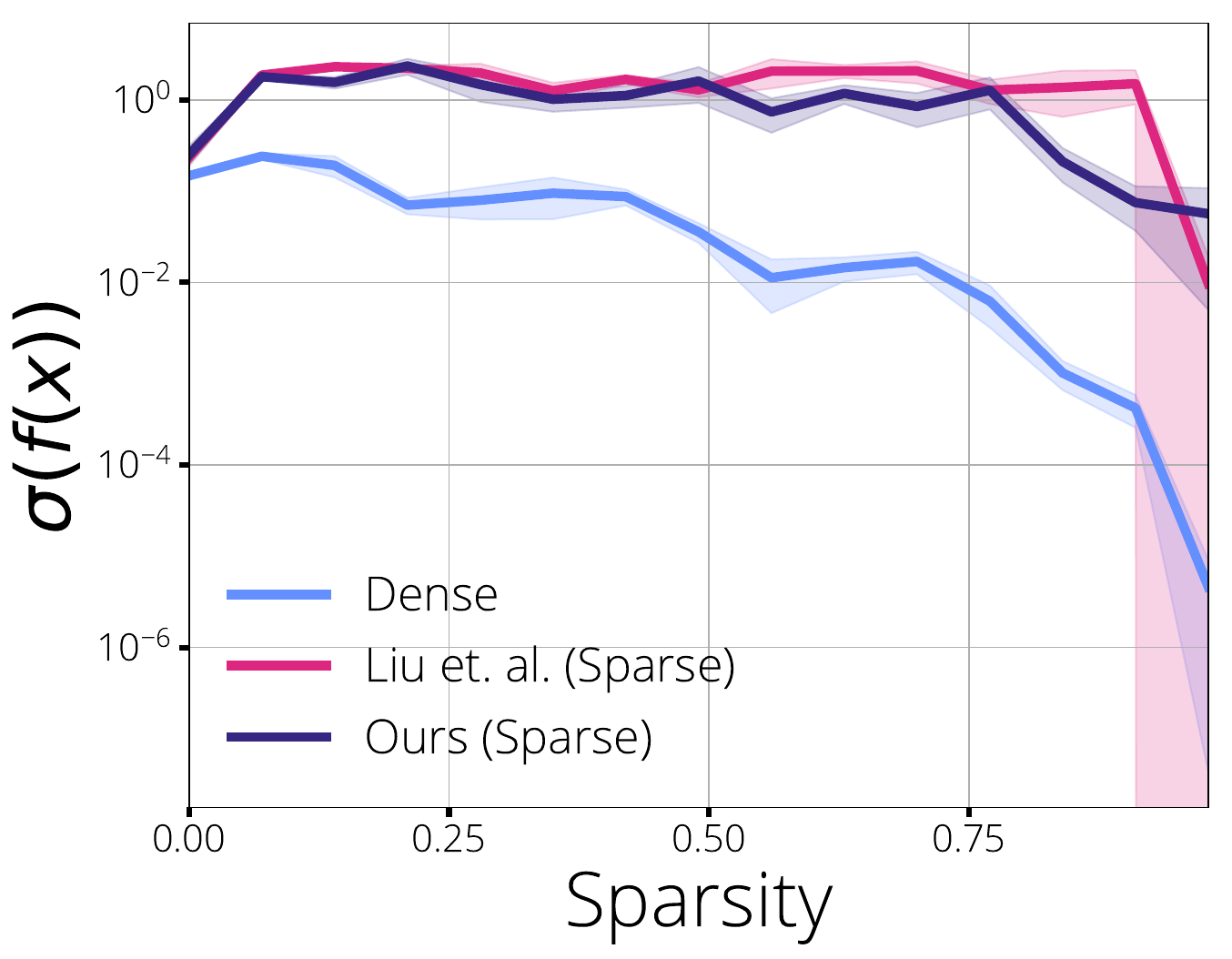}
\caption{Signal Propagation at Init.}\label{fig:initialization_signalprop}
\end{subfigure}%
\caption{\textbf{Glorot/He Initialization for a Sparse \gls{nn}.}
All neurons in a dense \gls{nn} layer (\subref{fig:initialization_dense}) have the same fan-in, whereas in a sparse \gls{nn} (\subref{fig:initialization_sparse}) the fan-in can differ for \emph{every neuron}, potentially requiring sampling from a different distribution for every neuron. 
The initialization derivation/fan-out variant are explained further in \cref{initialization-full-explanation}.
(\subref{fig:initialization_signalprop}) Std.\ dev.\ of the pre-softmax output of LeNet5 with input  sampled from a normal distribution, over 5 different randomly-initialized sparse \gls{nn} for a range of sparsities.}%
\label{fig:initialization}
\end{figure*}

\subsection{The Initialization Problem in Sparse NNs}
\label{sec:methods_init}
Here we analyze the gradient flow at initialization for random sparse \glspl{nn}, motivating the derivation of a more general initialization for \gls{nn} with heterogeneous connectivity, such as in unstructured sparse \glspl{nn}.

In practice, without a method such as \gls{batchnorm}~\citep{Ioffe2015}, using the correct initialization can be the difference between being able to train a \gls{dnn}, or not --- as observed for VGG16 in our results (\cref{sec:exp_init}, \cref{table:results}). The initializations proposed by \citet{Glorot2010understanding,He2015delvingdeep} ensure that the output distribution of every neuron in a layer is zero-mean and of unit variance by sampling initial weights from a Gaussian distribution with a variance based on the number of incoming/outgoing connections for all the neurons in a dense layer, as illustrated in \cref{fig:initialization_dense}, which is assumed to be identical for all neurons in the layer.

In an unstructured sparse \gls{nn} however, the number of incoming/outgoing connections is not identical for all neurons in a layer, as illustrated in \cref{fig:initialization_sparse}.
In \cref{initialization-derivation} we derive the initialization for this more general case. Here we will focus only on explaining the generalized \citet{He2015delvingdeep} initialization for forward propagation, which we used in our experiments. In \cref{initialization-full-explanation} we explain in full the generalized \citet{Glorot2010understanding,He2015delvingdeep} initialization, in the forward, backward and average use cases. 

We propose to initialize every weight $w^{[\ell]}_{ij} \in W^{n^{[\ell]}\times n^{[\ell - 1]}}$ in a sparse layer $\ell$ with $n^{[\ell]}$ neurons, and connectivity mask $[m_{ij}^{[\ell]}] = M^{\ell} \in [0, 1]^{n^{[\ell]}\times n^{[\ell - 1]}} $ with,
\begin{equation}
        w_{ij}^{[\ell]} \sim  \mathcal{N}\left(0,\,\frac{2}{ \FanIn_i^{[\ell]}}\right),\label{eqn:sparseinit}
\end{equation}
where $\FanIn_i^{[\ell]} = \sum_{j=1}^{n^{[\ell-1]}} m_{ij}^{[\ell]}$ is the number of incoming connections for neuron $i$ in layer $\ell$.

In the special case of a dense layer where $m_{ij}^{[\ell]} = 1,\,\forall i, j$, \cref{eqn:sparseinit} reduces to the initialization proposed by \citet{He2015delvingdeep} since $\FanIn_i^{[\ell]} = n^{[\ell - 1]},\, \forall i$. Similarly, the initialization proposed by \citet{Liu2018} is another special case where it is assumed $\FanIn_i^{[\ell]} \equiv \E_k[\FanIn_k^{[\ell]}]$, i.e.\ all neurons have the same number of incoming connections in a layer which is often not true with unstructured sparsity. Using the dense initialization in a sparse \gls{dnn} causes signal to vanish, as empirically observed in \cref{fig:initialization_signalprop}, whereas sparsity-aware initialization techniques (ours and \citet{Liu2018}) keep the variance of the signal constant.

\subsection{Dynamic Sparse Training}%
\label{sec:methods_dst}
While initialization is important for the first training step, the gradient flow during the early stages of training is not well addressed by initialization alone, rather it has been addressed in dense \glspl{dnn} by normalization methods~\citep{Ioffe2015}. Our findings show that even with \gls{batchnorm} however, the gradient flow during the training of unstructured sparse \glspl{nn} is poor.

Recently, a promising new approach to training sparse \glspl{nn} has emerged --- \Acrfull{dst} --- that learns connectivity adaptively during training, showing significant improvements over baseline methods that use fixed masks. These methods perform periodic updates on the sparse connectivity of each layer: commonly replacing least magnitude connections with new connections selected using various criteria. We consider two of these methods and measure their effect on gradient flow during training: \gls{set}~\citep{Mocanu2018}, which chooses new connections randomly and \gls{rigl}~\citep{Evci2019}, which chooses the connections with high gradient magnitude. Understanding why and how these methods achieve better results can help us in improving upon them. 

\subsection{Lottery Ticket Hypothesis}
\label{sec:methods_lottery}
A recent approach for training unstructured sparse \glspl{nn} while achieving similar generalization to the original dense solution is the \gls{lth}~\citep{Frankle2018lottery}. Notably, rather than training a pruned \gls{nn} structure from random initialization, the \gls{lth} uses the dense initialization from which the pruning solution was derived.

\paragraph{Definition [\Acrlong{lth}]:} Given a \gls{nn} architecture $f$ with parameters $\theta$ and an optimization function $O^N(f,\theta)$ = $\theta^N$, which gives the optimized parameters of $f$ after $N$ training steps, there exists a sparse sub-network characterized by the binary mask $M$ such that for some iteration $K$, $O^N(f,\theta^K*M)$ performs as well as $O^N(f,\theta)*M$, whereas the model trained from another random initialization $\theta_S$, using the same mask $O^N(f,\theta_S*M)$, typically does not\footnote[1]{See \citet{Frankle2019stabilizing} for details. $*$ indicates element-wise multiplication, respecting the mask.}.
Initial results of \citet{Frankle2018lottery} showed the \gls{lth} held for $K=0$, but later results ~\citep{Liu2018,Frankle2019stabilizing} showed a larger $K$ is necessary for larger datasets and \gls{nn} architectures, i.e.\ $N\gg{}K\ge{}0$.

\begin{table*}[tp]
\small
\centering
 \begin{adjustbox}{max width=\textwidth}
\begin{threeparttable}
\caption{\textbf{Results of Trained Sparse/Dense Models from Different Initializations.}
The initialization proposed in \cref{eqn:sparseinit} (Ours) and \citet{Liu2018} improve generalization consistently over masked dense (Original) except for in ResNet50. Note that VGG16 trained without a sparsity-aware initialization fails to converge in some instances. \emph{Baseline} corresponds to the original dense architecture, whereas \emph{Small Dense} corresponds to a smaller dense model with approx.\ the same parameter count as the sparse models.}%
\label{table:results}
\begin{tabular}{@{}p{2em}rrrrrrrrr@{}}
\toprule
& \multicolumn{3}{c}{MNIST} & \multicolumn{6}{c}{\imgnet} \\
\cmidrule(lr){2-4} \cmidrule(lr){5-10}
& \multicolumn{3}{c}{LeNet5 (95\% sparse)} & \multicolumn{3}{c}{VGG16 (80\% sparse)} & \multicolumn{3}{c}{ResNet50 (80\% sparse)}\\
\cmidrule(lr){2-4} \cmidrule(lr){5-7} \cmidrule(lr){8-10}
Baseline & \multicolumn{3}{c}{99.21$\pm$0.07} & \multicolumn{3}{c}{69.25$\pm$0.13} & \multicolumn{3}{c}{76.75$\pm$0.12}\\
Lottery & \multicolumn{3}{c}{98.26$\pm$0.27} & \multicolumn{3}{c}{0.10$\pm$0.01} & \multicolumn{3}{c}{75.75$\pm$0.12{\tnote{*}}}\\
Small Dense & \multicolumn{3}{c}{98.21$\pm$0.46} & \multicolumn{3}{c}{61.75$\pm$0.09} & \multicolumn{3}{c}{71.95$\pm$0.24}\\[1.2ex]
& \multicolumn{1}{c}{Dense} & \multicolumn{1}{c}{Sparse (Liu)} & \multicolumn{1}{c}{Sparse (Ours)} & \multicolumn{1}{c}{Dense} & \multicolumn{1}{c}{Sparse (Liu)} & \multicolumn{1}{c}{Sparse(Ours)} & \multicolumn{1}{c}{Dense} & \multicolumn{1}{c}{Sparse (Liu)} & \multicolumn{1}{c}{Sparse (Ours)}\\
\cmidrule(lr){2-2} \cmidrule(lr){3-3} \cmidrule(lr){4-4} \cmidrule(lr){5-5} \cmidrule(lr){6-6} \cmidrule(l){7-7}  \cmidrule(l){8-8}  \cmidrule(l){9-9}  \cmidrule(l){10-10}
Scratch & 62.99$\pm$42.16  & 96.64$\pm$0.83 & {\bf 97.70}$\pm$0.09    & 51.81$\pm$3.02    & {\bf 62.71}$\pm$0.05 & 62.52$\pm$0.10  & 70.58$\pm$0.18
& {\bf 70.72}$\pm$0.16 & 70.63$\pm$0.22\\
\acrshort{set} & 63.33$\pm$42.44 &  97.77$\pm$0.31& {\bf 98.16}$\pm$0.06   & 53.55$\pm$1.03   & {\bf 63.19}$\pm$0.26 & 63.13$\pm$0.15  & {\bf 72.93}$\pm$0.27 & 72.77$\pm$0.27 & 72.56$\pm$0.14\\
\acrshort{rigl} &  80.82$\pm$34.74 & \textbf{98.14}$\pm$0.17 & {\bf 98.13}$\pm$0.09    & 37.15$\pm$26.20    & {\bf 63.69}$\pm$0.02 &  63.56$\pm$0.06  & {\bf 74.41}$\pm$0.05 & 74.38$\pm$0.10 & 74.38$\pm$0.01\\ 
\bottomrule
\end{tabular}
\begin{tablenotes}
\footnotesize
\item[*] With late-rewinding (i.e.\ $K=5000$).
\end{tablenotes}
\end{threeparttable}
\end{adjustbox}
\vspace{-1em}
\end{table*}
We measure the gradient flow of \glspl{lt} and observe poor gradient flow overall. Despite this, \glspl{lt} enjoy significantly faster convergence compared to regular \gls{nn} training, which motivates our investigation of the success of \glspl{lt} beyond gradient flow. 
\Glspl{lt} require the connectivity mask as found by the pruning solution along with parameter values from the early part of the dense training. 
Given the importance of the early phase of training~\citep{Frankle2020early,lewkowycz2020large}, it is natural to ask about the difference between \glspl{lt} and the solution they are derived from (i.e.\ pruning solutions). Answering this question can help us understand if the success of \glspl{lt} is primarily due to its relation to the solution, or if we can identify generalizable characteristics that help with sparse \glspl{nn} training.
\begin{figure*}[h]
  \centering
  \begin{subfigure}{.33\textwidth}
  \centering
  \includegraphics[width=\linewidth]{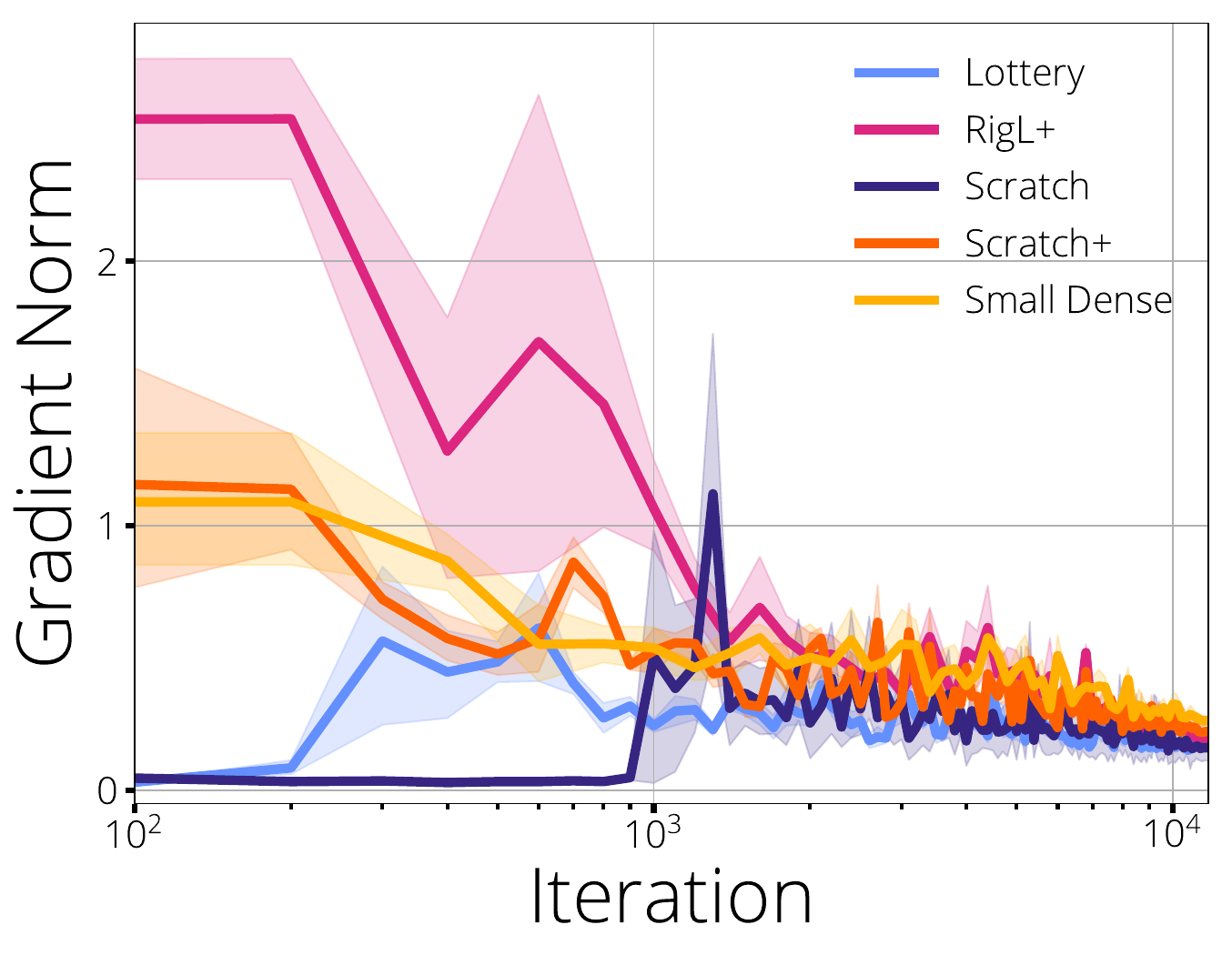}
  \caption{LeNet5}%
\label{subfig:gradnorm1}
\end{subfigure}%
    \begin{subfigure}{.33\textwidth}
  \centering
  \includegraphics[width=\linewidth]{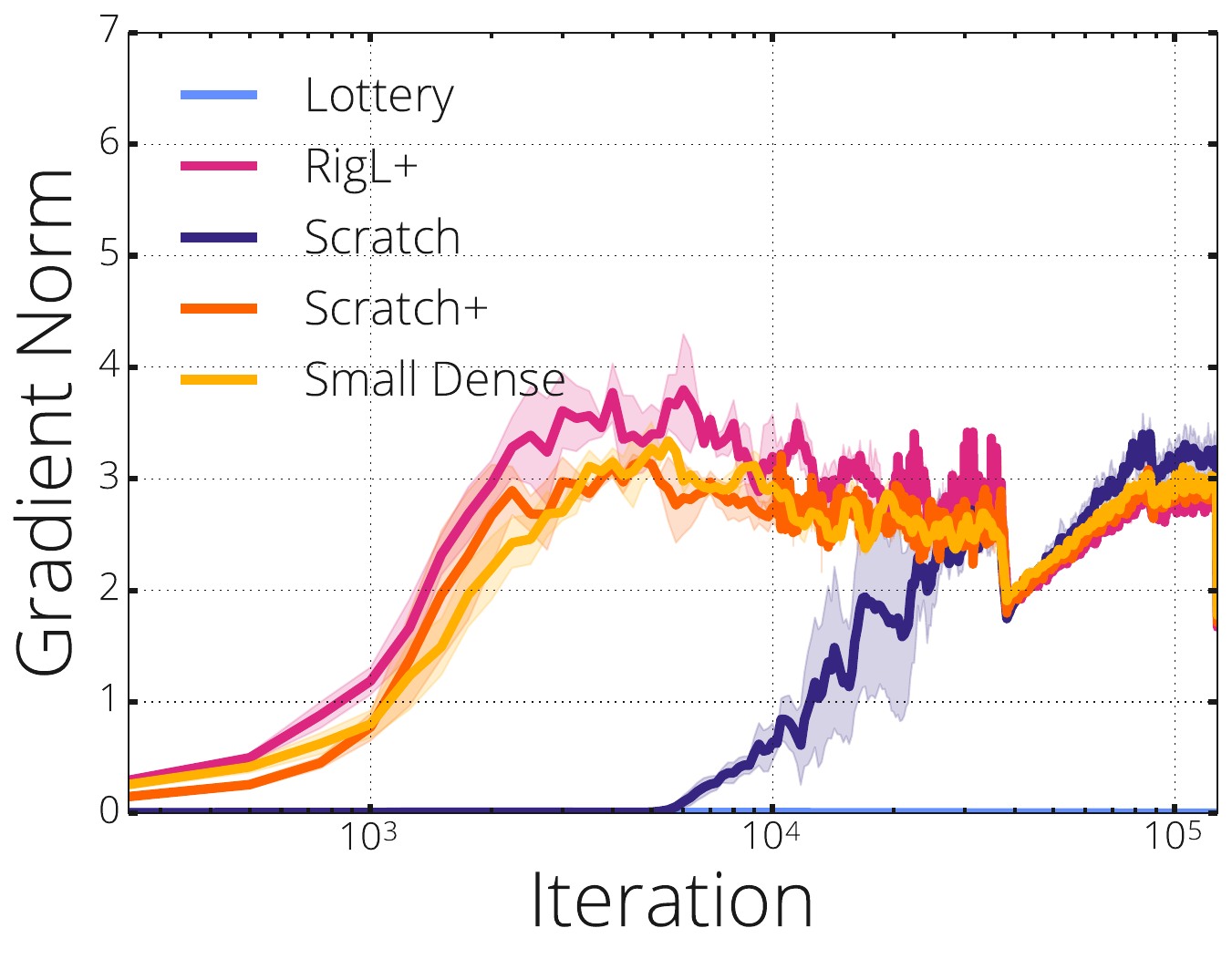}
    \caption{VGG-16}%
\label{subfig:gradnorm2}
\end{subfigure}%
\begin{subfigure}{.33\textwidth}
  \centering
  \includegraphics[width=\linewidth]{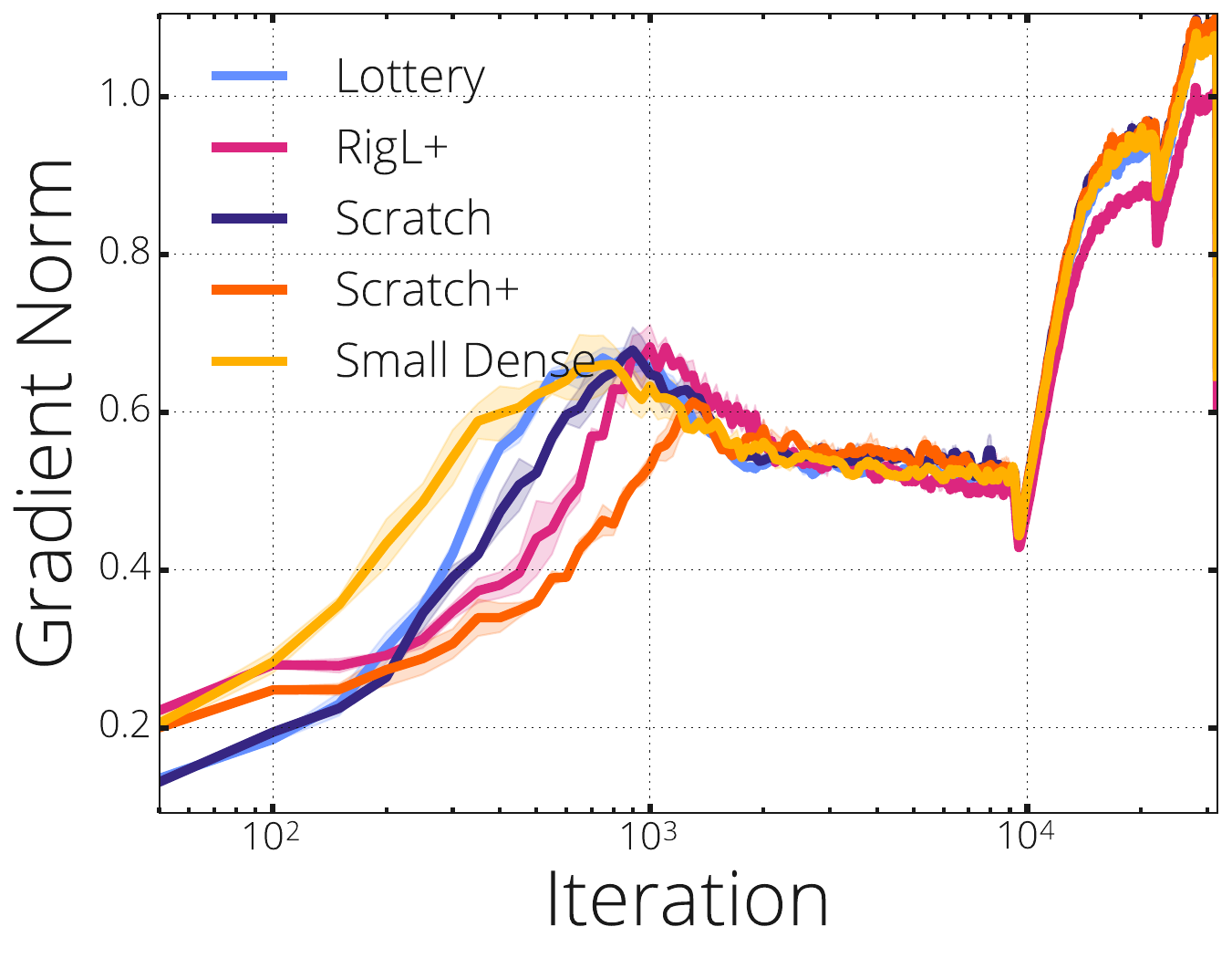}
    \caption{ResNet-50}%
\label{subfig:gradnorm3}
\end{subfigure}%
    \caption{\textbf{Gradient Flow of Sparse Models during Training.} Gradient flow during training averaged over multiple runs, `+` indicates training runs with our proposed sparse initialization and Small Dense corresponds to training of a dense network with same number of parameters as the sparse networks. Lottery ticket runs for ResNet-50 include late-rewinding.}
    \label{fig:gradnorm}
\vspace{-1em}
\end{figure*}%

%

\section{Experiments}
\label{sec:exp}
Here we show empirically that
(1) sparsity-aware initialization improves gradient flow at initialization for all methods, and achieves higher generalization for networks without \gls{batchnorm},
(2) the mask updates of \gls{dst} methods increase gradient flow and create new negative eigenvalues in the Hessian; which we believe to be the main factor for improved generalization,
(3) lottery tickets have poor gradient flow, however they achieve good performance by effectively re-learning the pruning solution, meaning they do not address the problem of training sparse \glspl{nn} in general.
Our experiments include the following settings: LeNet5 on MNIST, VGG16 on \imgnet~ and ResNet-50 on \imgnet{}. Experimental details can be found in \cref{sec:experimentaldetails}\footnote{Implementation of our sparse initialization, Hessian calculation and code for reproducing our experiments can be found at \url{https://github.com/google-research/rigl/tree/master/rigl/rigl_tf2}. Additionally we provide videos that shows the evolution of Hessian during training under different algorithms in the supplementary material.}.
\subsection{Gradient Flow at Initialization}
\label{sec:exp_init}
In this section, we measure the gradient flow over the course of the training (\cref{fig:gradnorm}) and evaluate the performance of generalized He initialization method (\cref{table:results}), and that proposed by \citet{Liu2018}, over the commonly used dense initialization in sparse \gls{nn}. Additional gradient flow plots for the remaining methods are shared in \cref{sec:gradnorm_app}.
Sparse \glspl{nn} initialized using the initialization distribution of a dense model (\textit{Scratch} in \cref{fig:gradnorm}) start in a flat region where gradient flow is very small and thus initial progress is limited. Learning starts after 1000 iterations for LeNet5 and 5000 for VGG-16, however, generalization is sub-optimal. 
\begin{figure*}[tp]
\centering
\begin{subfigure}[t]{.45\linewidth}
\includegraphics[width=\linewidth]{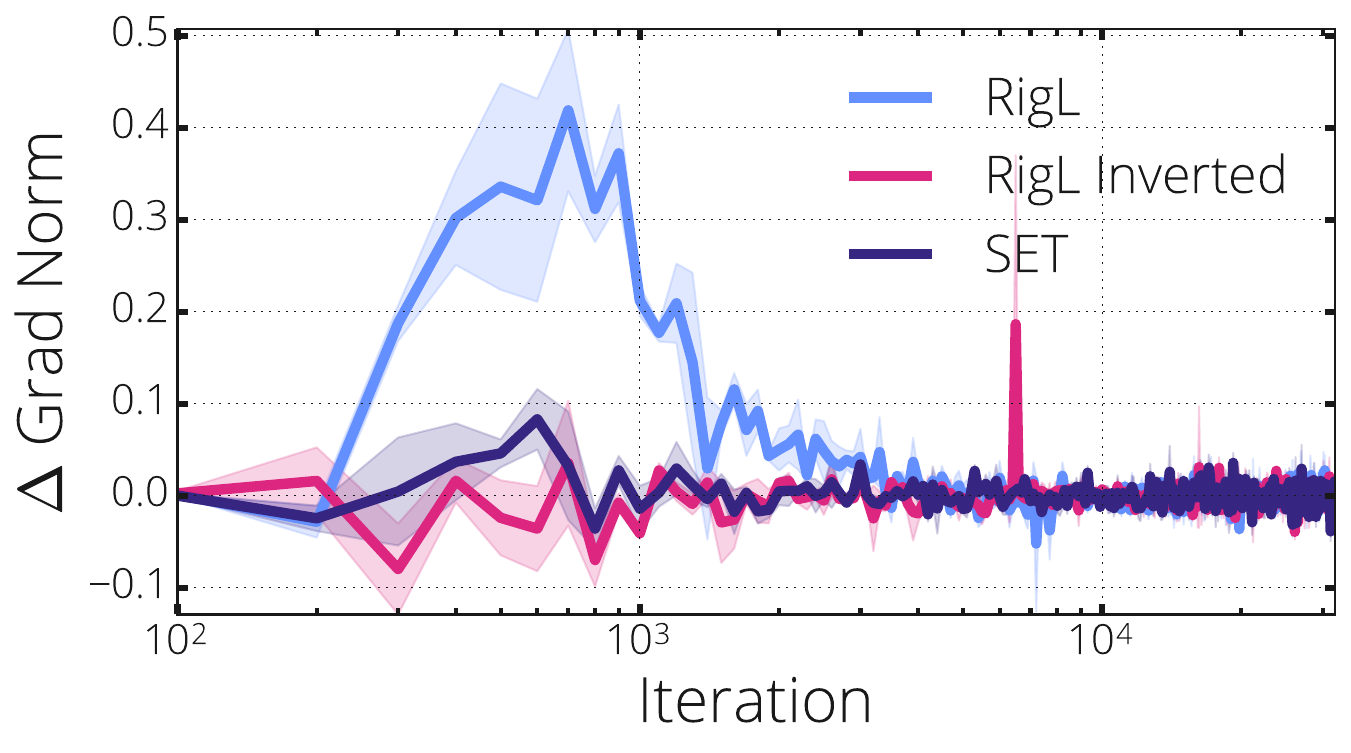}
\caption{ResNet-50}%
\label{subfig:dynamicsparse1}
\end{subfigure}%
\begin{subfigure}[t]{.45\linewidth}
\includegraphics[width=\linewidth]{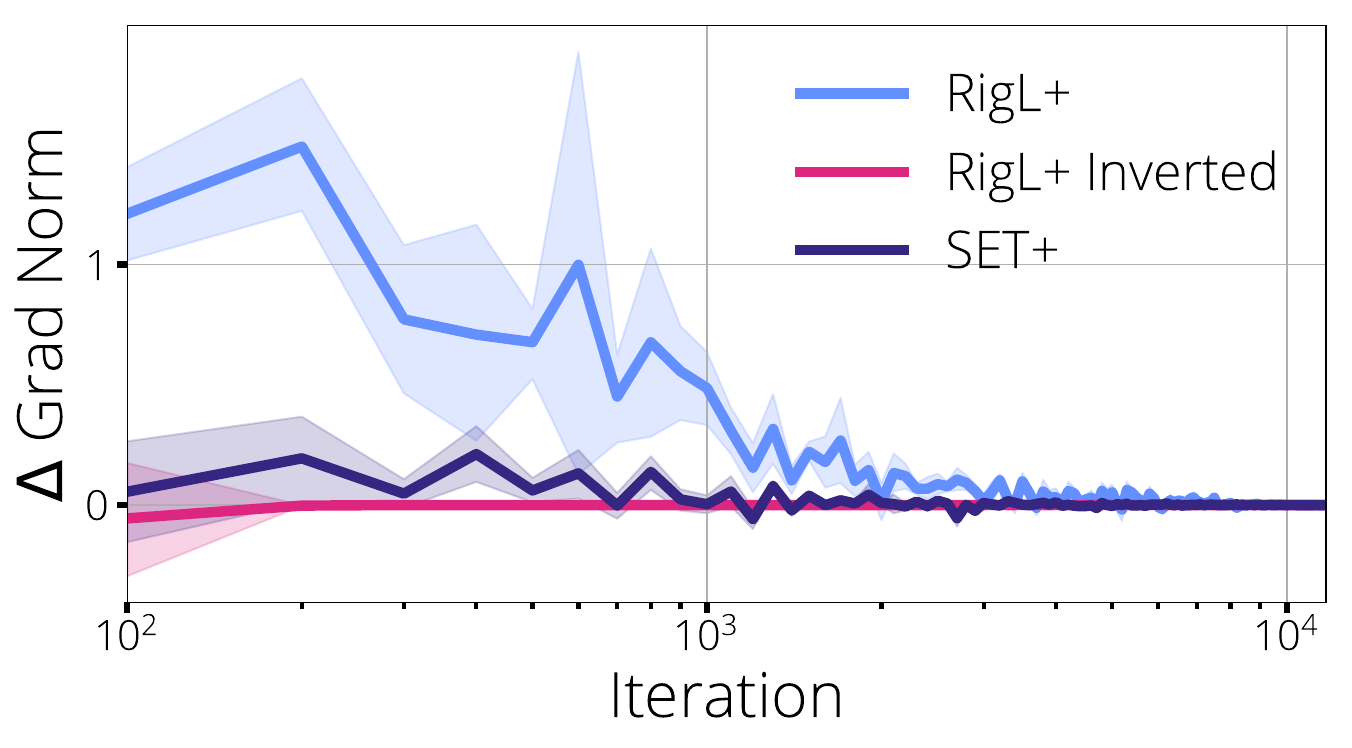}
\caption{LeNet5}%
\label{subfig:dynamicsparse2}
\end{subfigure}%

\caption{\textbf{Effect of Mask Updates in \acrlong{dst}.} Effect of mask updates on the gradient norm. \textit{RigL Inverted} chooses connections with least magnitude. We measure the gradient norm before and after the mask updates and plot the $\Delta$. `+` indicates proposed initialization and used in MNIST experiments.}\label{fig:dynamicsparse}
\vspace{-1em}
\end{figure*}
\Citet{Liu2018} reported their proposed initialization has \emph{no empirical effect} as compared to the masked dense initialization\footnote{Models with \gls{batchnorm} and skip connections are less affected by initialization, and this is likely why the authors did not observe this effect.}. In contrast, our results show their method to be largely as effective as our proposed initialization, despite the \citet{Liu2018} initialization being incorrect in the general case (see \cref{sec:methods_init}). This indicates that the assumption of a neuron having roughly uniform connectivity is sufficient for the ranges of sparsity considered, possibly due to a law-of-large-numbers-like averaging effect. We expect this effect to disappear with higher sparsity and our initialization to be more important. Results in \cref{tab:lenet98} for a 98\% sparse LeNet5 demonstrate this, in which our sparsity-aware initialization provides 40\% improvement on test accuracy. Our initialization remedy the vanishing gradient problem at initialization (\textit{Scratch+} in \cref{fig:gradnorm}) and result in better generalization for all methods. For instance, improved initialization results in an 11\% improvement in Top-1 accuracy for VGG16 (62.52 vs 51.81).
\begin{table}[t]
\begin{tabular}{@{}lccc@{}}
\toprule
        & Dense/Other & Sparse (Liu) & Sparse (Ours)\\
\cmidrule{2-4}
Scratch & 11.35$\pm$0.00   & 56.51$\pm$36.90           & \textbf{90.00$\pm$4.04}         \\
RigL    & 11.35$\pm$0.00 & 62.34$\pm$41.63         & \textbf{96.72$\pm$0.23}\\
\cmidrule{2-4}\
Prune   & 96.50$\pm$0.31  & -- & -- \\ \bottomrule
\end{tabular}
\caption{\textbf{Sparsity-aware Initialization and Generalization: Sparse (98\%) LeNet5 (MNIST)}. Results show the importance of our sparsity-aware initialization at higher sparsity regimes, compared with a standard dense initialization~\citep{He2015delvingdeep}, or scaled-sparse initialization~\citep{Liu2018}.}
\label{tab:lenet98}
\end{table}
While initialization is extremely important for \glspl{nn} without \gls{batchnorm} and skip connections, its effect on modern architectures, such as Resnet-50, is limited~\citep{Evci2019,fixup,Frankle2020pruningatinit}. We confirm these observations in our ResNet-50 experiments in which, despite some initial improvement in gradient flow, sparsity-aware initializations seem to have no effect on final generalization. Additionally, we observe significant increase in gradient norm after each learning rate drop (due to increased gradient variance), which suggests studying the gradient norm in the latter part of the training might not be helpful.

The \gls{lt} hypothesis holds for MNIST and ResNet50 (when K=5000) but not for VGG16. We observe poor gradient flow for \glspl{lt} at initialization similar to \textit{Scratch}. After around 2000 steps gradients become non zero for \textit{Scratch}, while gradient flow for \gls{lt} experiments stay constant. Our sparse initialization improves gradient flow at initialization and we observe a significant difference in gradient flow during early training between RigL and Scratch+. \gls{dst} methods alone can help resolve gradient flow problem at initialization and thus bring better performance. This is reflected in \cref{table:results}: Scratch, SET and RigL obtains 63\%, 63\% and 81\% respectively; all starting from the bad initialization (Dense).

\subsection{Gradient Flow During Sparse Training}
\label{sec:exp_dst}
Our hypothesis for \cref{fig:gradnorm} is that the \gls{dst} methods improve gradient flow through the updates they make on the sparse connectivity, which in turn results in better performance. To verify our hypothesis, we measure the change in gradient flow and Hessian spectrum whenever the sparse connectivity is updated. We also run the inverted baseline for RigL (\textit{RigL Inverted}), in which the growing criteria is reversed and connections with the \emph{smallest} gradients are activated as in \citet{Frankle2020pruningatinit}.

\Gls{dst} methods such as \gls{rigl} replace low saliency connections during training. Assuming the pruned connections indeed have a low impact on the loss, we might expect to see increased gradient norm after new connections are activated, especially in the case of \gls{rigl}, which picks new connections with high magnitude gradients. In \cref{fig:dynamicsparse} we confirm that \gls{rigl} updates increase the norm of the gradient significantly, especially in the first half of training, whereas \gls{set}, which picks new connections randomly, seems to be less effective at this. Using the inverted \gls{rigl} criteria doesn't improve the gradient flow, as expected, and without this \gls{rigl}'s performance degrades (73.83$\pm$0.12 for ResNet-50 and 92.71$\pm$7.67 for LeNet5). These results suggest that improving gradient flow early in training --- as \gls{rigl} does --- might be the key for training sparse networks. Various recent works that improve \gls{dst} methods support this hypothesis: longer update intervals \citet{Liu2021DoWA} and adding parallel dense components \citep{Price2021DenseFT} or activations \citep{curci2021} that improve gradient flow brings better results. Additional plots (\cref{sec:gradnorm_app}) using different initialization methods demonstrate the same behaviour. 

\begin{figure*}[t]
  \centering
  
  \includegraphics[width=0.75\linewidth]{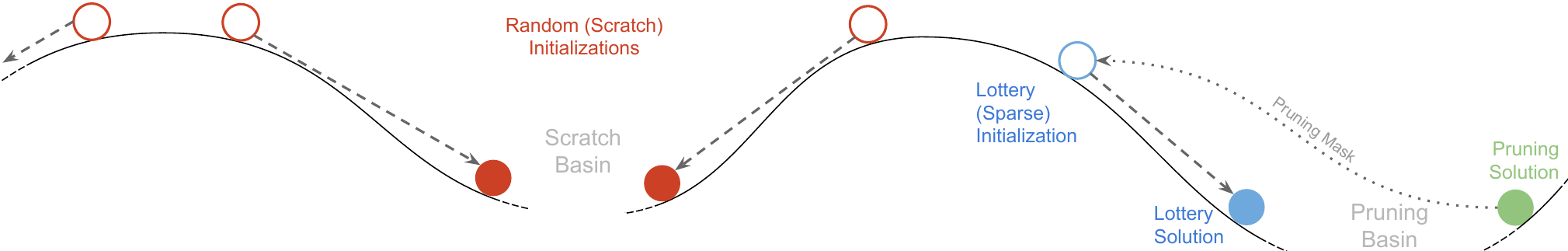}
  \caption{\textbf{\Acrlongpl{lt} Are Biased Towards the Pruning Solution, Unlike Random Initialization.} A cartoon illustration of the loss landscape of a sparse model, after it is pruned from a dense solution to create a \gls{lt} sub-network.
  A \gls{lt} initialization is within the basin of attraction of the pruned model's solution.
  In contrast a random initialization is unlikely to be close to the dense solution's basin.}%
  \label{fig:lotteryticketsexplained}
\vspace{-1em}
\end{figure*}%

When the gradient is zero, or uninformative due to the error term of the approximation, analyzing the Hessian could provide additional insights~\citep{sagun2017, ghorbani2019, papyan2019}. In \cref{sec:lenet5hessian}, we show the Hessian spectrum before and after sparse connectivity updates. After \gls{rigl} updates we observe more negative eigenvalues with significantly larger magnitudes as compared to \gls{set}. Large negative eigenvalues can further accelerate optimization if corresponding eigenvectors are aligned with the gradient. We leave verifying this hypothesis as a future work.

\subsection{Why Lottery Tickets are Successful}
\label{sec:exp_lottery}

We found that  \glspl{lt} do not improve gradient flow, either at initialization, or early in training, as shown in \cref{fig:gradnorm}. This may be surprising given the apparent success of \glspl{lt}, however the questions posed in \cref{sec:methods_lottery} present an alternative hypothesis for the ease of training from a \gls{lt} initialization: Can the success of \glspl{lt} be due to their relationship to well-performing pruning solutions? Here we present results showing that indeed (1) \glspl{lt} initializations are consistently closer to the pruning solution than a random initialization, (2) trained \glspl{lt} (i.e.\ \gls{lt} solutions) consistently end up in the same basin as the pruning solution and (3), \gls{lt} solutions are highly similar to pruning solutions under various function similarity measures. Our resulting understanding of \glspl{lt} in the context of the pruning solution and the loss landscape is illustrated in \cref{fig:lotteryticketsexplained}. 

\begin{figure*}[tp]
\centering
\begin{subfigure}[b]{.28\textwidth}
\centering
\includegraphics[width=\linewidth]{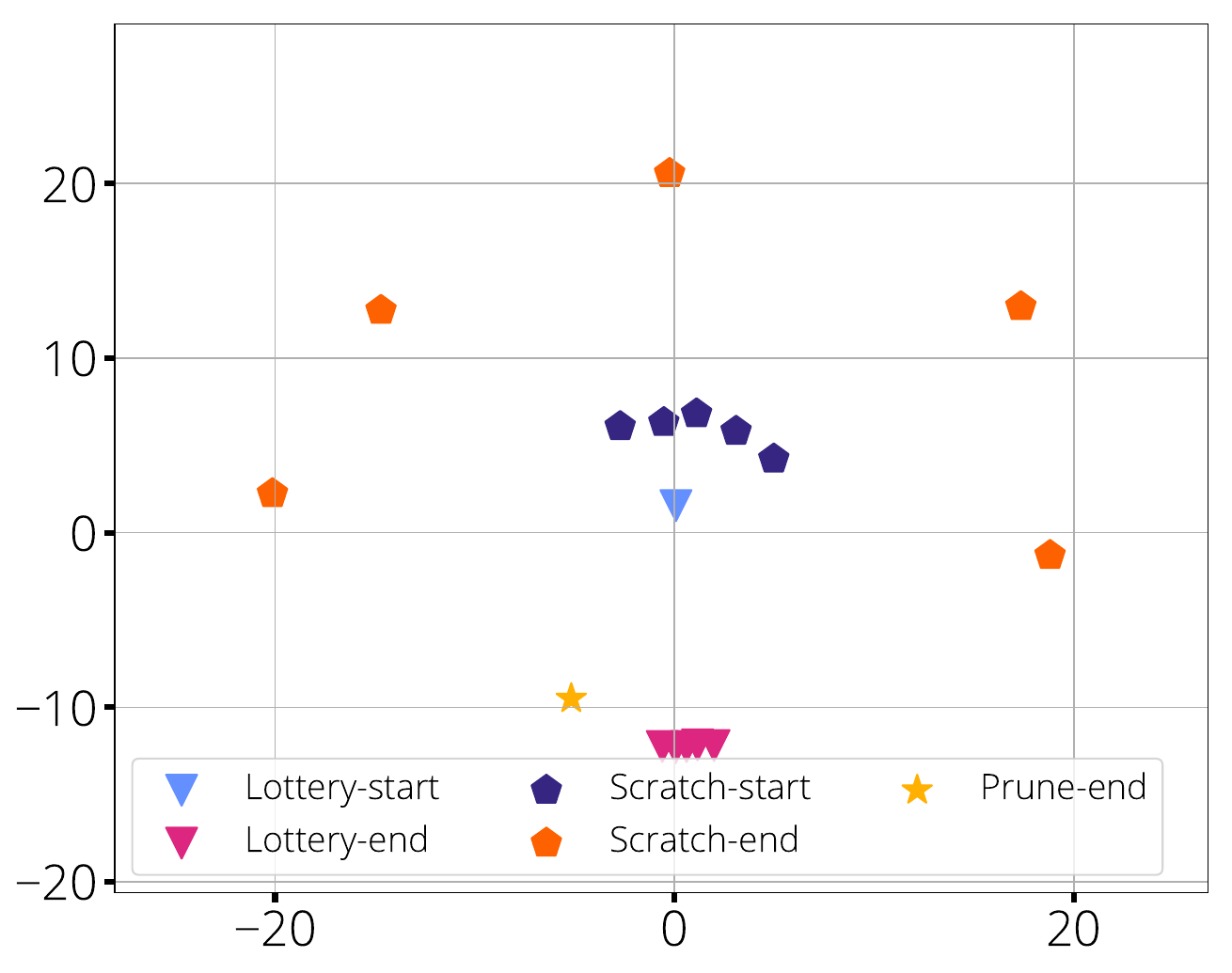}
\caption{\acrshort{mds} Projection}\label{subfig:mnistmds}
\end{subfigure}%
~
\begin{subfigure}[b]{.3\textwidth}
\centering
\textbf{MNIST/LeNet5}\\
\vspace{1em}
{
\footnotesize
\begin{tabular}{@{}p{2.4em}p{2.5em}p{3em}@{}}
\toprule
Method & LT & Scratch\\
\midrule
$Acc_{test}$    & 98.52 & 97.19\\
$d_{init}$     & 13.61 & 17.46\\
$d_{final}$       & 8.03 & 25.64\\ 
\bottomrule
\end{tabular}
}
\vspace{2.5em} 
\caption{L2 Distances}\label{subfig:mnistl2table}
\end{subfigure}%
~
\begin{subfigure}[b]{.3\textwidth}
\centering
\includegraphics[width=\linewidth]{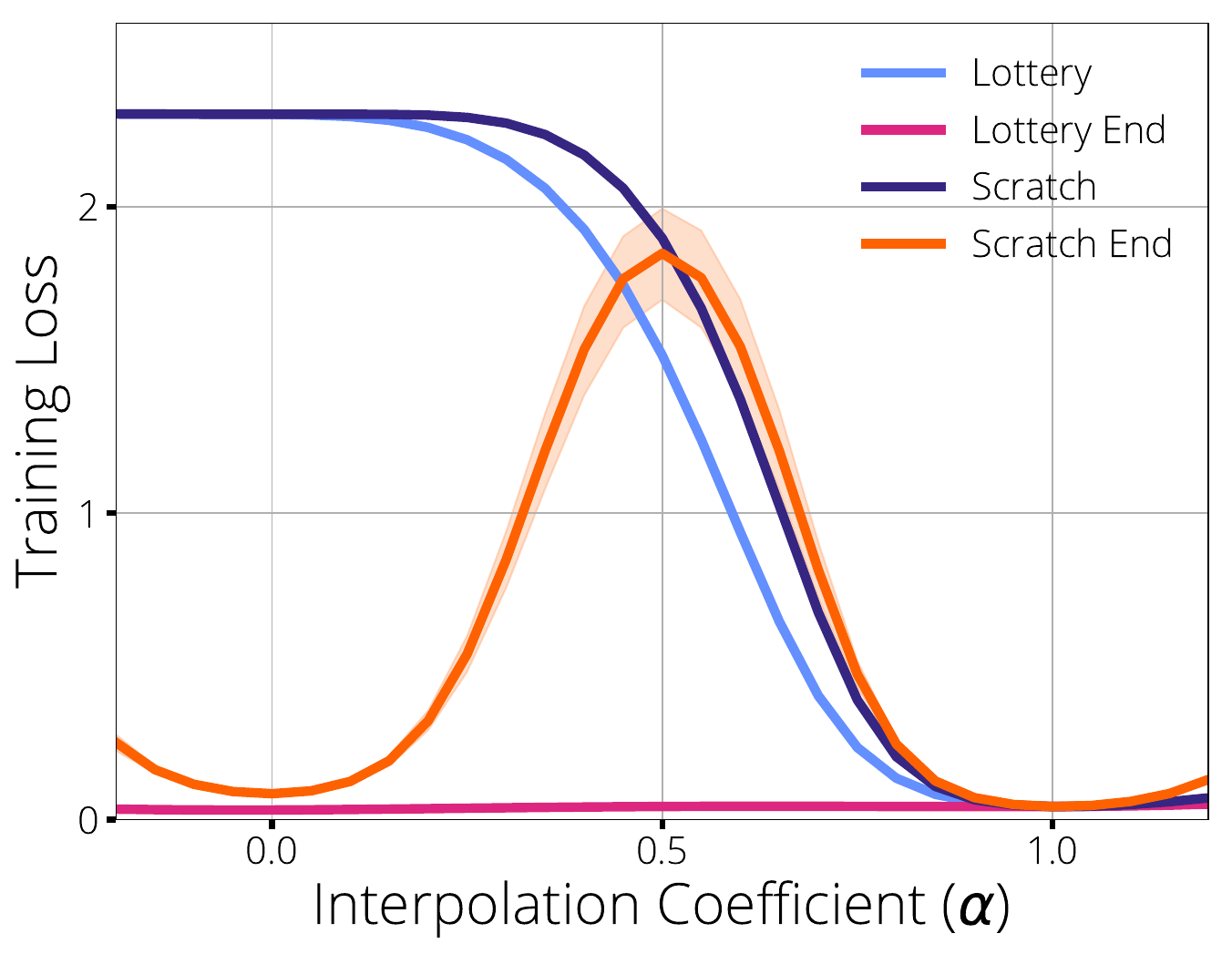}
\caption{Loss Interpolation}\label{subfig:mnistinterpolation}
\end{subfigure}\\
\begin{subfigure}[b]{.28\textwidth}
\centering
\includegraphics[width=\linewidth]{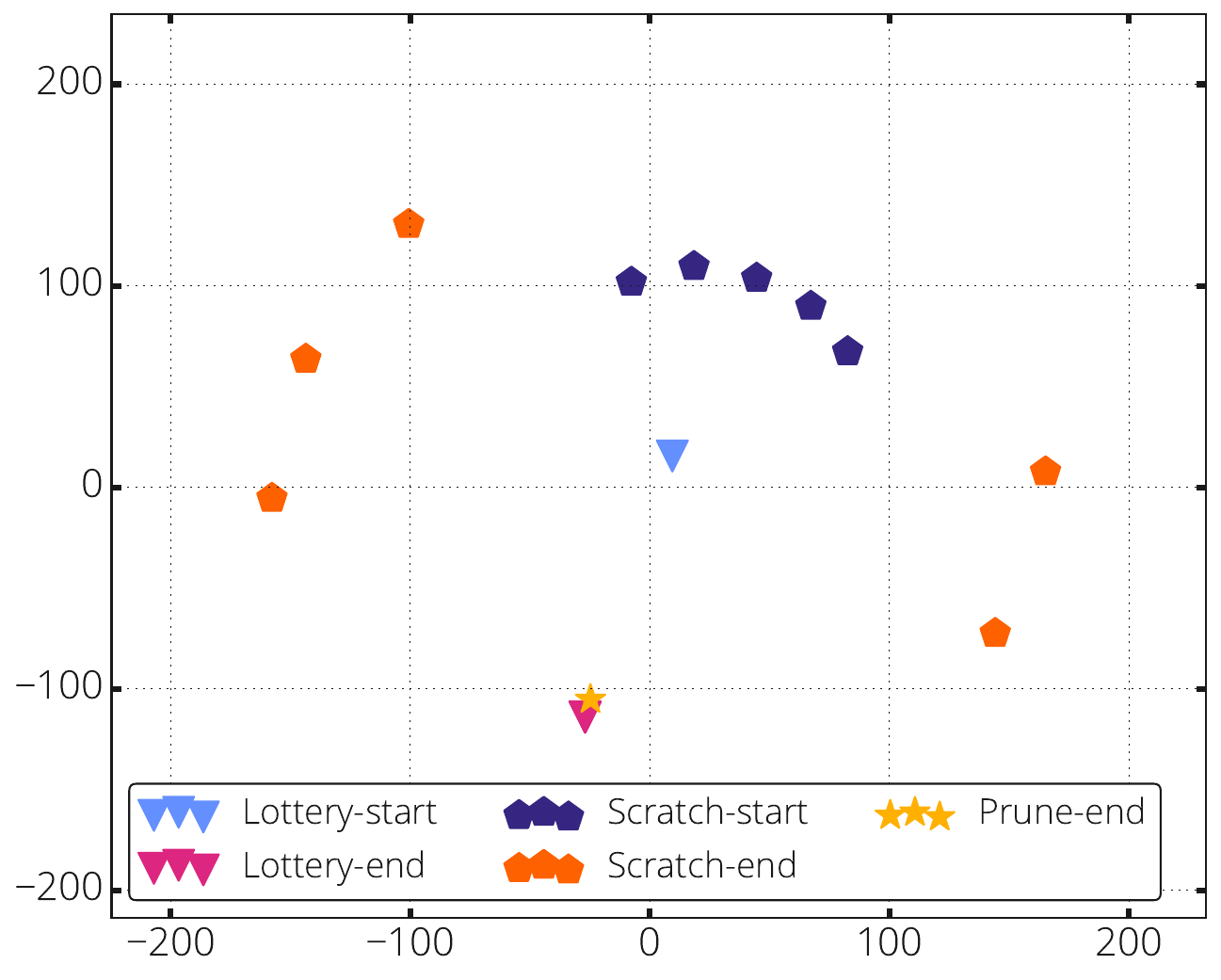}%
\caption{\Acrshort{mds} Projection}%
\label{subfig:imagenetmds}
\end{subfigure}%
~
\begin{subfigure}[b]{.3\textwidth}
\centering
\textbf{\imgnet{}\\
/ResNet-50
}\\
\vspace{1em}
{
\footnotesize
\begin{tabular}{@{}p{2.4em}p{2.5em}p{3em}@{}}
\toprule
Method          & LT & Scratch   \\
\midrule
$Acc_{test}$    & 75.74& 71.16 \\ 
$d_{init}$     & 147.12& 200.44        \\ 
$d_{final}$       & 39.35 & 215.98        \\ 
\bottomrule
\end{tabular}%
}
\vspace{1.5em} 
\caption{L2 Distances}%
\label{subfig:imagenetl2table}
\end{subfigure}%
~
\begin{subfigure}[b]{.3\textwidth}
\centering
\includegraphics[width=\linewidth]{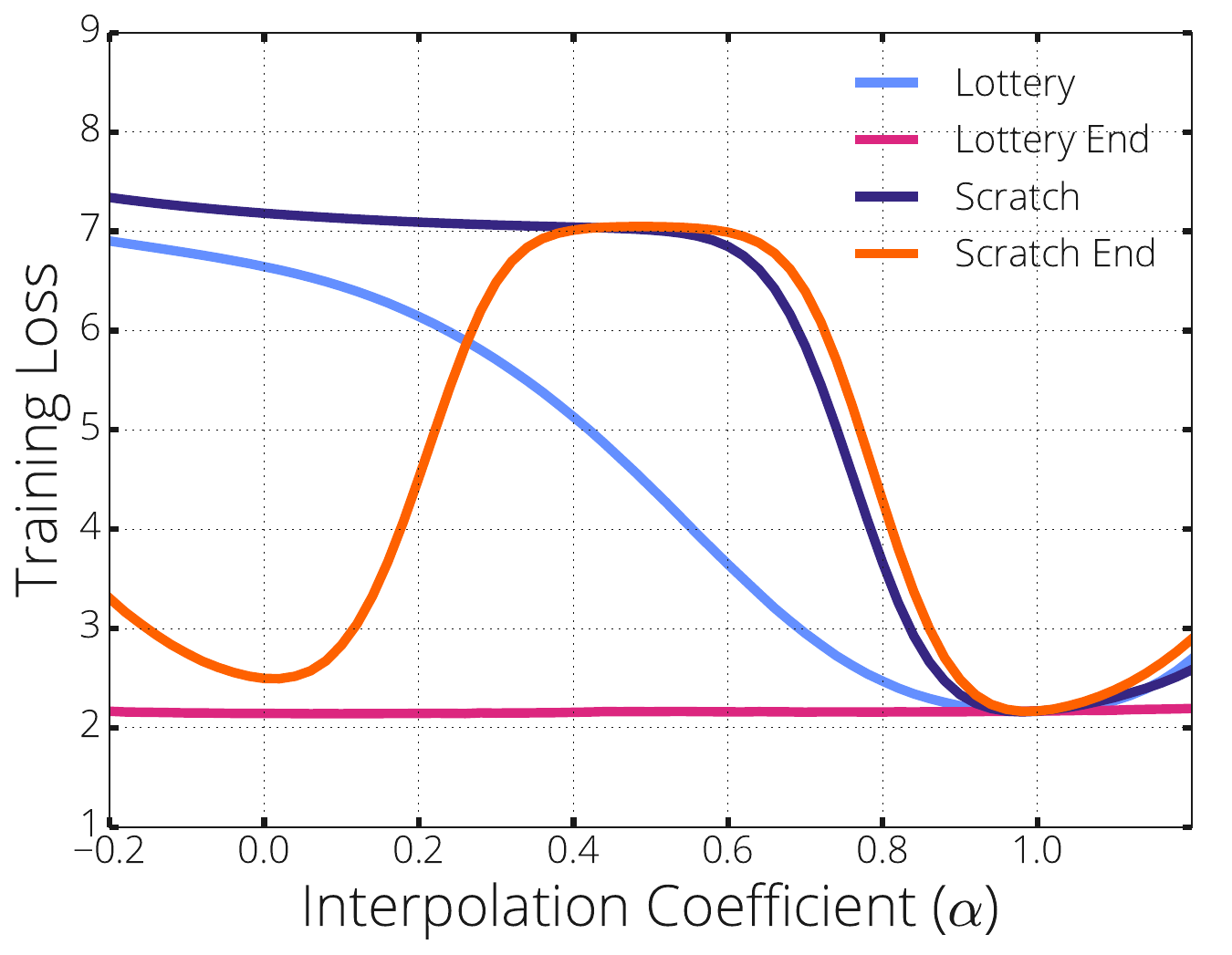}%
\caption{Loss Interpolation}%
\label{subfig:imagenetinterpolation}
\end{subfigure}%
\caption{\textbf{\Acrshort{mds} Embeddings/L2 Distances}: (\subref{subfig:mnistmds}, \subref{subfig:imagenetmds}): 2D \gls{mds} embedding of sparse \glspl{nn} with the same connectivity/mask; (\subref{subfig:mnistl2table}, \subref{subfig:imagenetl2table}): the average L2-distance between a pruning solution and other derived sparse networks; (\subref{subfig:mnistinterpolation}, \subref{subfig:imagenetinterpolation}): linear path between the pruning solution ($\alpha=1.0$) and \gls{lt}/scratch at both initialization, and solution (end of training). Top and bottom rows are for MNIST/LeNet5 and \imgnet{}/ResNet-50 respectively.}%
\label{fig-sec2}
\vspace{-1em}

\end{figure*}
%
\paragraph{Experimental Setup} To investigate the relationship between the pruned and \gls{lt} solutions we perform experiments on two models/datasets: a 95\% sparse LeNet5\footnote{Note: We use ReLU activation functions, unlike the original architecture~\citep{lecun1989backpropagation}.} architecture~\citep{lecun1989backpropagation} trained on MNIST (where the original \gls{lt} formulation works, i.e.\ $K=0$), and an 80\% sparse ResNet-50~\citep{resnet} on \imgnet~\citep{imagenet} 
(where $K=0$ doesn't work~\citep{Frankle2019stabilizing}), for which we use values from $K=2000$ ($\approx$\engordnumber{6} epoch). In both cases, we find a \gls{lt} initialization by pruning each layer of a dense \gls{nn} separately using magnitude-based iterative pruning ~\citep{gupta2018}. Further details of our experiments are in \cref{sec:experimentaldetails}.
\paragraph{\Acrlongpl{lt} Are Close to the Pruning Solution} We train 5 different models using different seeds from both \emph{scratch} (random) and \gls{lt} initializations, the results of which are in \cref{subfig:mnistl2table,subfig:imagenetl2table}. These networks share the same pruning mask and therefore lie in the same solution space. We visualize distances between initial and final points of these experiments in \cref{subfig:mnistmds,subfig:imagenetmds} using 2D \acrfull{mds}~\citep{Kruskal1964mds} embeddings.  
\textbf{\textit{LeNet5/MNIST}}: In \cref{subfig:mnistl2table}, we provide the average L2 distance to the pruning solution at initialization ($d_{init}$), and after training ($d_{final}$). We observe that \gls{lt} initializations start significantly closer to the pruning solution on average ($d_{init}=13.61$ v.s.\ $17.46$). After training, \glspl{lt} end up more than 3$\times$ closer to the pruning solution compared to scratch.
\textbf{\textit{Resnet-50/\imgnet:}} We observe similar results for Resnet-50/\imgnet. \Glspl{lt}, again, start closer to the pruning solution, and solutions are 5$\times$ closer ($d_{final}=39.35$ v.s.\ $215.98$). These results explain non-random initial loss values for \gls{lt} initializations~\citep{Zhou2019deconstructing} and their inability to find good solutions if repelled by pruning solutions~\citep{maene2021}. \Glspl{lt} are biased towards the pruning solution they are derived from, but are they in the same basin? Can it be the case that \glspl{lt} learn significantly different solutions each time in function space despite being linearly connected to the pruning solution?
\paragraph{\Acrlongpl{lt} are in the Pruning Solution Basin} Investigating paths between different solutions is a popular tool for understanding how various points in parameter space relate to each other in the loss landscape~\citep{Goodfellow2014,Garipov2018,Draxler2018,Evci2019,Fort2020deep, Frankle2019linearmode}. For example, \citet{Frankle2019stabilizing} use linear interpolations to show that \glspl{lt} always go to the same basin\footnote{ We define a basin as a set of points, each of which is linearly connected to at least one other point in the set.} when trained in different data orders. In \cref{subfig:mnistinterpolation,subfig:imagenetinterpolation} we look at the linear paths between pruning solution and 4 other points: \gls{lt} initialization/solution and random (scratch) initialization/solution. Each experiment is repeated 5 times with different random seeds, and mean values are provided with 80\% confidence intervals. In both experiments we observe that the linear path between \gls{lt} initialization and the pruning solution decreases faster compared to the path that originates from scratch initialization. After training, the linear paths towards the pruning solution change drastically. The path from the scratch solution depicts a loss barrier; the scratch solution seems to be in a different basin than the pruning solution\footnote{This is not always true, it is possible that non-linear low energy paths exist between two solutions~\citep{Draxler2018,Garipov2018}, but searching for such paths is outside the scope of this work.}. In contrast, \glspl{lt} are linearly connected to the pruning solution in both small and large-scale experiments indicating that \glspl{lt} have the same basin of attraction as the pruning solutions they are derived from. While it seems likely, these results do not however explicitly show that the \gls{lt} and pruning solutions have learned similar functions. 

\begin{table*}[tp]

\small
\centering
\begin{threeparttable}
\caption{\textbf{Ensemble \& Prediction Disagreement}. We compare the function similarity~\citep{Fort2020deep} with the original pruning solution and ensemble generalization over 5 sparse models, trained from random initializations and \glspl{lt}. As a baseline, we also show results for 5 pruned models trained from different random initializations.
See \cref{sec:disagreementfull} for full results.}\label{disagreement}
\begin{tabular}{@{}llS[table-format=2.2(2)]S[table-format=2.2]S[table-format=1.4(4)]S[table-format=1.4(4)]@{}}
\toprule
&{Initialization} & {(Top-1) Test Acc.} & {Ensemble} & {Disagreement} & {Disagree.\ w/ Pruned}\\
\cmidrule{2-6}
\multirow{6}{3em}{LeNet5 MNIST} & LT & 98.52 \pm 0.02 & 98.58 & 0.0043\pm 0.0006 & 0.0089 \pm 0.0002\\ 
&Scratch & 97.04 \pm 0.15 & 98.00 & 0.0316\pm 0.0023 & 0.0278 \pm 0.0020\\
&Scratch (Diff. Init.)& 97.19 \pm 0.33 & 98.43 & 0.0352\pm 0.0037 & 0.0278 \pm 0.0032\\
&Prune Restart & 98.60 \pm 0.01 & 98.63 & 0.0027\pm 0.0003 & 0.0077 \pm 0.0003\\
\cmidrule(l){2-6}
&Pruned Soln. & 98.53 & {--} & {--} & {--}\\
&5 Diff.\ Pruned & 98.30 \pm 0.23 & 99.07 & 0.0214\pm 0.0023 & 0.0197 \pm 0.0019{\tnote{*}}\\
\cmidrule(l){1-6}
\multirow{4}{3em}{ResNet50 ImageNet} & LT & 75.73\pm0.08 & 76.27 & 0.0894\pm0.0009 & 0.0941\pm0.0009\\ 
&Scratch & 71.16\pm0.13 & 74.05 & 0.2039\pm0.0013 & .2033\pm0.0012\\ 
\cmidrule(l){2-6}
&Pruned Soln. & 75.60 & {--} & {--} & {--}\\
&5 Diff.\ Pruned & 75.65\pm0.13& 77.80 & 0.162\pm0.0008 & 0.1623\pm0.0011{\tnote{*}}\\
\bottomrule
\end{tabular}
\begin{tablenotes}
\footnotesize
\item[*] Here we compare 4 different pruned models with the pruning solution LT/Scratch are derived from.
\end{tablenotes}
\end{threeparttable}
\end{table*}%

\paragraph{\Acrlongpl{lt} Learn Similar Functions to the Pruning Solution}%
\Citet{Fort2020deep} motivate deep ensembles by empirically showing that models starting from different random initializations typically learn different solutions, as compared to models trained from similar initializations, and thus improve performance.
In \cref{sec:disagreementfull} we adopt the analysis of \citep{Fort2020deep}, but in comparing \gls{lt} initializations and random initializations using \emph{fractional disagreement} --- the fraction of class predictions over which the LT and scratch models disagree with the pruning solution they were derived from, \gls{kl}, and \gls{jsd}. The results in \cref{disagreementfull} suggest \glspl{lt} models converge on a solution almost identical to the pruning solution and therefore an ensemble of \glspl{lt} brings marginal improvements. Our results also show that having a fixed initialization alone can not explain the low disagreement observed for \gls{lt} experiments as \textit{Scratch} solutions obtain an average disagreement of $0.0316$ despite using the same initialization, which is almost 10 times larger than that of the \gls{lt} solutions ($0.0043$). As a result, \gls{lt} ensembles show significantly less gains on MNIST and \imgnet~($+0.06\%$ and $+0.54\%$ respectively) compared to a random sparse initialization ($+0.96\%$ and $+2.89\%$).
\paragraph{Implications:}
\textbf{(a) Rewinding of \glspl{lt}.}~\citet{Frankle2019stabilizing,Frankle2019linearmode} argued that \glspl{lt} work when the training is \textit{stable}, and thus converges to the same basin when trained with different data sampling orders. In \cref{sec:exp_lottery}, we show that this basin is the same one found by pruning, and since the training converges to the same basin as before, we expect to see limited gains from rewinding if any. This is partially confirmed by \citet{renda2020} which shows that restarting the learning rate schedule from the pruning solution performs better than rewinding the weights. \textbf{(b) Transfer of \glspl{lt}.}~Given the close relationship between \glspl{lt} and pruning solutions, the observation that \glspl{lt} trained on large datasets transfer to smaller ones, but not \emph{vice versa} \citep{Morcos2019oneticket,Sabatelli2020transferability} can be explained by a common observation in transfer learning: networks trained in large datasets transfer to smaller ones. \textbf{(c) \gls{lt}'s Robustness to Perturbations.}~\Citet{Zhou2019deconstructing,Frankle2020early} found that certain perturbations, like only using the signs of weights at initialization, do not impact \gls{lt} generalization, while others, like shuffling the weights, do. Our results bring further insights to these observations: As long as the perturbation is small enough such that a \gls{lt} stays in the same basin of attraction, results will be as good as the pruning solution. \textbf{(d) Success of \glspl{lt}.}~While it is exciting to see widespread applicability of \glspl{lt} in different domains~\citep{Brix2020successfully, Li2020towards, venkatesh2020}, the results presented in this paper suggest this success may be due to the underlying pruning algorithm (and transfer learning) rather than \gls{lt} initializations themselves.
%
%
\section{Conclusion}
\label{sec:conclusion}

In this work we studied (1) why training unstructured sparse networks from random initialization performs poorly and; (2) what makes \glspl{lt} and \gls{dst} the exceptions? 
We identified that randomly initialized unstructured sparse \glspl{nn} exhibit poor gradient flow when initialized naively and proposed an alternative initialization that scales the initial variance for each neuron separately. Furthermore we showed that modern sparse \gls{nn} architectures are more sensitive to poor gradient flow during early training rather than initialization alone. We observed that this is somewhat addressed by state-of-the-art \gls{dst} methods, such as \gls{rigl}, which significantly improves gradient flow during early training over traditional sparse training methods. 
Finally, we show that \glspl{lt} do not improve gradient flow at either initialization or during training, but rather their success lies in effectively re-learning the original pruning solution they are derived from. We showed that a \glspl{lt} initialization resides within the same basin of attraction as the pruning solution and, furthermore, when trained the \gls{lt} solution learns a highly similar solution to the pruning solution, limiting their ensemble performance. These findings suggest that \glspl{lt} are fundamentally limited in their potential for improving the training of sparse \glspl{nn} more generally.
\section*{Acknowledgements}
We like to thank members of the Google Brain team for their useful feedback and support. Specifically, we thank Sara Hooker, Fabian Pedregosa, Laura Graesser, Ilya Tolstikhin, Erich Elsen, Ross Goroshin and Danny Tarlow for providing feedback on the preprint. We thank Ying Xiao for helping out calculating hessian of large networks.

\bibliography{main}
\newpage
\appendix

\renewcommand*{\thefootnote}{\fnsymbol{footnote}}
\section{Experimental Details}\label{sec:experimentaldetails}
\begin{table*}[tp]
\centering
\footnotesize
\begin{threeparttable}
\caption{\textbf{\cref{sec:exp_lottery}: Experiment Details/Hyperparameters}. Initial Learning Rate (LR), LR Schedule (Sched.), Batchsize (Batch.), Momentum ($m$), Weight Decay (WD), $t_\textrm{start}$, $t_{\textrm{end}}$ and $f$ are the pruning starting iteration, end iteration, and mask update frequency respectively.}\label{table:section2hyperparameters}
\begin{tabular}{@{}p{3em}p{3em}lrrrp{2em}p{1em}p{1.2em}rrrr@{}}
\toprule
Dataset & Model & $t_{\textrm{total}}$ & Epochs & Batch. & LR & Sched. & $m$ & WD & Sparsity & \multicolumn{3}{c}{Pruning}\\
& & & & & & & & & & $t_{\textrm{start}}$ & $t_{\textrm{end}}$ & $f$\\
\midrule
\multirow{1}{*}{MNIST} & \multirow{1}{*}{LeNet5} & 11719 & \multirow{1}{*}{30} & \multirow{1}{*}{128} & \multirow{1}{*}{0.05} & \multirow{1}{*}{Cosine} & \multirow{1}{*}{0.9} & \multirow{1}{*}{0} &\multirow{1}{*}{95\%} & 3000 & 7000 & 100\\
\cmidrule(l){1-13}
\multirow{1}{*}{ImageNet} & \multirow{1}{*}{ResNet50} & 32000 & \multirow{1}{*}{$\approx$102} & 4096 & \multirow{1}{*}{1.6} & Step\tnote{*}& \multirow{1}{*}{0.9} & \multirow{1}{*}{$1\times 10^{-4}$} & \multirow{1}{*}{80\%} & 5000 & 8000 & 2000\\
\bottomrule
\end{tabular}
\begin{tablenotes}
\footnotesize
\item[*] Step schedule has a linear warm-up in first 5 epochs and decreases the learning rate by a factor of $10$ at epochs $30, 70$ and $90$.
\end{tablenotes}
\end{threeparttable}
\end{table*}

\begin{table*}[tp]
\centering
\footnotesize
\begin{threeparttable}
\caption{\textbf{\cref{sec:exp_init}: Experiment Details/Hyperparameters}. Initial Learning Rate (LR), LR Schedule (Sched.), Batchsize (Batch.), Momentum ($m$), Weight Decay (WD), Initial Drop Fraction (Drop.), $t_{\textrm{end}}$ and $f$ are the pruning mask update frequency and end iteration respectively. \textit{LeNet5+} row corresponds the LeNet5 experiments with our sparse initialization, whereas \textit{LeNet5} is the regular masked initialization. }\label{table:section4hyperparameters}
\begin{tabular}{@{}p{3em}p{3em}lrrrp{2em}p{0.9em}p{1.8em}p{1.4em}rrr@{}}
\toprule
Dataset & Model & $t_{\textrm{total}}$ & Epochs & Batch. & LR & Sched. & $m$ & WD & Sparsity & \multicolumn{3}{c}{\Acrshort{dst}}\\
& & & & & & & & & & \multirow{1}{2em}{Drop.} & $f$ & $t_{\textrm{end}}$\\
\midrule
\multirow{1}{*}{MNIST} & \multirow{1}{*}{LeNet5} & \multirow{1}{*}{11719} & \multirow{1}{*}{30} & \multirow{1}{*}{128} & \multirow{1}{*}{0.05} & \multirow{1}{*}{Cosine} & \multirow{1}{*}{0.9} & \multirow{1}{*}{\tiny$5\times10^{-4}$} & \multirow{1}{*}{95\%} & 0.3 &\multirow{1}{*}{500} & \multirow{1}{*}{11719}\\
\cmidrule(l){1-13}
\multirow{3}{*}{ImageNet} & \multirow{1}{*}{ResNet50} & 32000 & \multirow{3}{*}{$\approx$102} & 4096 & 1.6 & \multirow{3}{*}{Step\tnote{*}} & \multirow{3}{*}{0.9} & \multirow{3}{*}{\tiny{$1\times10^{-4}$}} & \multirow{3}{*}{80\%} & 0.3 &100 & \multirow{3}{*}{25000}\\
\cmidrule(l){2-3} \cmidrule(lr){5-6} \cmidrule(lr){11-12} 
& \multirow{1}{*}{VGG16} & 128000 & & 1024 & 0.04 & & & & &0.1&500&\\
\bottomrule
\end{tabular}
\begin{tablenotes}
\footnotesize
\item[*] Step schedule has a linear warm-up in first 5 epochs and decreases the learning rate by a factor of $10$ at epochs $30, 70$ and $90$.
\end{tablenotes}
\end{threeparttable}
\end{table*}
\subsection{Details of Experiments in Section \ref{sec:exp_lottery}}
The training hyper-parameters used in \cref{sec:exp_lottery} are shared in \cref{table:section2hyperparameters}. All experiments in this section start with a pruning experiment, after which the sparsity masks found by pruning are used to perform \gls{lt} experiments. We use iterative magnitude pruning~\citep{gupta2018} in our experiments, which is a well studied and more efficient pruning method as compared to the one used by \citet{Frankle2018lottery}. Our pruning algorithm performs iterative pruning without rewinding the weights between intermediate steps and requires significantly less iterations. We expect our results would be even more pronounced with additional rewinding steps.

We use SGD with momentum in all of our experiments. \textit{Scratch} and \textit{Lottery} experiments use the same hyper-parameters. Additional specific details of our experiments are shared below.

\paragraph{LeNet5} We prune all layers of LeNet5, so that they reach 95\% final sparsity (i.e.\ 95\% of the parameters are zeros). We choose this sparsity, since at this sparsity, we start observing stark differences between \textit{Lottery} and \textit{Scratch} in terms of performance. We set the weight decay to zero, similar to the MNIST experiments done in the original \gls{lt} paper~\citep{Frankle2018lottery} and do a grid search over learning-rates=\{0.1,0.2,0.05,0.02,0.01\}. Loss values for the linear interpolation experiments are calculated on the entire training set.

\paragraph{ResNet50} We prune all layers of ResNet50, except the first layer, so that they reach 80\% final sparsity. In this setting rewinding to the original initialization doesn't work, hence we use values from \engordnumber{6} epoch. Loss values for the linear interpolation experiments are calculated using 500,000 images from the \imgnet{} training set.

\subsection{Details of Experiments in Section \ref{sec:exp_init} and \ref{sec:exp_dst}}
Training hyper-parameters used for these experiments are shared in \cref{table:section4hyperparameters}. 

\paragraph{MNIST} In this setting, the hyper-parameters are almost same as in \cref{sec:exp_lottery}, except we enable weight decay as it brings better generalization. We do a grid search over weight-decays=\{0.001,0.0001,0.00005,0.00001,0.0005\} and learning-rates=\{0.1,0.2,0.05,0.02,0.01\} and pick the values with top test accuracy. We use the masks found by pruning experiments in all of our MNIST experiments in this section to isolate the effect of the initialization.  We simplify the update schedule of \Acrfull{dst} methods such that they decay with learning rate. This approach fits well, since the original decay function used in these experiments is the cosine decay which is the same as our learning rate schedule. We scale learning rate such that it matches the initial drop fraction provided. Mask update frequency and initial drop fraction are chosen from a grid search of \{50, 100, 500\} and \{0.01,0.1,0.3\} respectively. To allow fair comparison, we use Glorot scaling in all of our initializations (i.e. scale=1 and we average fan-in fan-out values) as it is the default initialization for Tensorflow layers and our results shows that it out-performs He initialization by a small margin with the hyper-parameters used. Using He initialization brings similar results. 

\paragraph{Hessian calculation} The Hessian is calculated on full training set using Hessian-vector products. We mask our network after each gradient call and calculate only non-zero rows. After calculating the full Hessian, we use \texttt{numpy.eigh}~\citep{walt2011numpy} to calculate eigenvalues of the Hessian. 

\paragraph{\imgnet} In this setting, hyper-parameters are almost the same as in \cref{sec:exp_lottery} except for VGG16 architecture, where we use a smaller batch size and learning rate. For all \gls{dst} methods, we use a cosine drop schedule~\cite{dettmers2019} and hyper-parameters proposed by \citet{Evci2019}. For VGG, we reduce the mask update frequency and the initial drop fraction, as we observe better performance after doing a grid search over \{50, 100, 500\} and \{0.1,0.3,0.5\} respectively. We also use a non-uniform (ERK) sparsity distribution among layers as described in \citet{evci2019b}, since we observed that it brings better performance. 
\section{Glorot/He Initialization Generalized to Neural Networks with Heterogeneous Connectivity: Full Explanation/Derivation}%
\begin{figure*}[tp]
\centering
\begin{subfigure}{.32\textwidth}
\includegraphics[width=\textwidth]{imgs/masked_initialization_dense}
\caption{Dense Layer}
\end{subfigure}%
~
\begin{subfigure}{.66\textwidth}
\includegraphics[width=\textwidth]{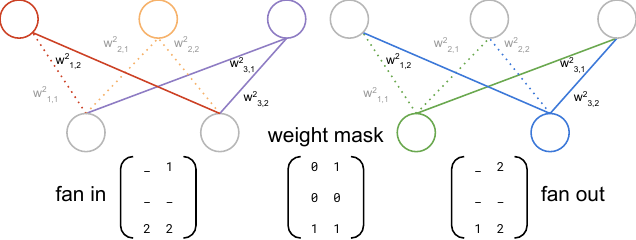}
\caption{Sparse Layer}
\end{subfigure}%
\caption{\textbf{Glorot/He Initialization for a Sparse \gls{nn}.}
\Citep{Glorot2010understanding,He2015delvingdeep} restrict the outputs of all neurons to be zero-mean and of unit variance. All neurons in a dense \gls{nn} layer (\subref{fig:initialization_dense}) have the same fan-in/fan-out, whereas in a sparse \gls{nn} (\subref{fig:initialization_sparse}) the fan-in/fan-out can differ for \emph{every neuron}, potentially requiring sampling from a different distribution for every neuron. The fan-in matrix contains the values used in \cref{eqn:sparseinit} for each neuron.}%
\label{fig:initialization_full}
\end{figure*}
Here we derive the full generalized initialization for both the forwards/backwards cases (i.e.\ fan-in/fan-out), refer to \cref{fig:initialization_full} for an illustration of how the connectivity for the fan-in/fan-out cases are determined for each neuron.
\subsection{Generalized Glorot/He Initialization: Backwards, Forwards and Average Cases}
\label{initialization-full-explanation}
For every weight $w^{[\ell]}_{ij} \in W^{n^{[\ell]}\times n^{[\ell - 1]}}$ in a layer $\ell$ with $n^{[\ell]}$ neurons, connecting neuron $i$ in layer $\ell$ to neuron $j$ in layer $(\ell-1)$ with $n^{[\ell - 1]}$ neurons, and weight mask $[m_{ij}^{[\ell]}] = M^{\ell} \in [0, 1]^{n^{[\ell]}\times n^{[\ell - 1]}} $,
\begin{align}
    \begin{array}{lc}
        \textrm{\small~\citet{Glorot2010understanding}:}& w_{ij}^{[\ell]} \sim  \mathcal{N}\left(0, \frac{1}{u}\right)\\
        \textrm{\small~\citet{He2015delvingdeep}:}& w_{ij}^{[\ell]} \sim  \mathcal{N}\left(0,\,\frac{2}{u}\right) \,
    \end{array}
\end{align}
\begin{align*}
    u =
    \begin{cases}
        \FanIn_i^{[\ell]} & \text{(forward)} \\
        \FanOut_j^{[\ell]} & \text{(backward)} \\
        \left(\FanIn_i^{[\ell]} + \FanOut_j^{[\ell]} \right)/2 & \text{(average)}\\
    \end{cases}\label{eqn:initializationfull}
\end{align*}
where,
$$ \FanIn_i^{[\ell]} = \sum_{j=1}^{n^{[\ell-1]}} m_{ij}^{[\ell]},\qquad\FanOut_j^{[\ell]} = \sum_{i=1}^{n^{[\ell]}} m_{ij}^{[\ell]},$$
are the number of incoming and outgoing connections respectively. In the special case of a dense layer where $m_{ij}^{[\ell]} = 1,\,\forall i, j$, \cref{eqn:sparseinit} reduces to the initializations proposed by~\citep{Glorot2010understanding,He2015delvingdeep} since $\FanIn^{[\ell]}_i = n^{[\ell-1]},\, \forall i$, and $\FanOut^{[\ell]}_j = n^{[\ell]},\, \forall j$.
\subsection{Derivation: Fixed Mask, Forward Propagation}
\label{initialization-derivation}
Given a sparse \gls{nn}, where the output of a neuron $a_{i}$ is given by,
$a_{i}^{[\ell]} = f\left(z_{i}^{[\ell]}\right)$, where $z_{i}^{[\ell]} = \sum_j^{n^{[\ell-1]}} m_{ij}^{[\ell]} w_{ij}^{[\ell]} a_{j}^{[\ell-1]},$
where $m_{ij}^{[\ell]} \in M^{[\ell]}$ and $w_{ij}^{[\ell]} \in W^{[\ell]}$ are the mask and weights respectively for layer $\ell$, and $a_{j}^{[\ell-1]}$ the output of the previous layer. Assume the mask $M^{[\ell]} \in \mathbbm{1}^{n^{[\ell]}\times n^{[\ell - 1]}}$ is constant, where $\mathbbm{1}^{n^{[\ell]}\times n^{[\ell - 1]}}$ is an indicator matrix.

As in \citet{Glorot2010understanding} we want to ensure $\Var(a_{i}^{[\ell]}) = \Var(a_{i}^{[\ell-1]})$, and $\mean(a_{i}^{[\ell]}) = 0$. Assume that $f(x) \approx x $ for $x$ close to $0$, e.g.\ in the case of $f(x)=\tanh(x)$, and that $w_{ij}^{[\ell]}$ and $a_{j}^{[\ell-1]}$ are independent,
\begin{align}
    \Var(a_{i}^{[\ell-1]})& \approx \Var(z_{i}^{[\ell]})&\\
    & = \Var\left(\sum_{j=1}^{n^{[l-1]}} m_{ij}^{[\ell]} w_{ij}^{[\ell]} a_{j}^{[\ell-1]}\right)&\\
    &= \sum_{j=1}^{n^{[\ell-1]}}\Var\left(m_{ij}^{[\ell]} w_{ij}^{[\ell]} a_{j}^{[\ell-1]}\right)\\
    & \hspace{5em} \textrm{(independent sum)}\\
    &= \sum_{j=1}^{n^{[\ell-1]}} \left(m_{ij}^{[\ell]}\right)^2 \Var\left(w_{ij}^{[\ell]} a_{j}^{[\ell-1]}\right)\\
    & \hspace{5em} \because m_{ij}^{[\ell]} \textrm{ is constant}, \Var(cX) = c^2\,\Var(X)\\
    &= \sum_{j=1}^{n^{[\ell-1]}}\; m_{ij}^{[\ell]}\; \Var\left(w_{ij}^{[\ell]} a_{j}^{[\ell-1]}\right)\\
    & \hspace{5em} \because m_{ij}^{[\ell]} \in [0, 1], \left(m_{ij}^{[\ell]}\right)^2 = m_{ij}^{[\ell]}.\\
    & = \sum_{j=1}^{n^{[\ell-1]}}\; m_{ij}^{[\ell]}\; \Var(w_{ij}^{[\ell]})\; \Var(a_{j}^{[\ell-1]}).\\
    & \hspace{5em} \textrm{(independent product)}\label{eqn:beforeweightassumption}
\end{align}
Assume $\Var(w_{im}^{[\ell]}) = \Var(w_{in}^{[\ell]}),\, \forall n,m$, i.e.\ the variance of all weights for a given neuron are the same, and $\Var(a_{n}^{[\ell-1]}) = \Var(a_{m}^{[\ell-1]})$, i.e.\ the variance of any of the outputs of the previous layer are the same. 
Therefore we can simplify \cref{eqn:beforeweightassumption}, 
\begin{align}
    \Var(a_{i}^{[\ell-1]})& = \sum_{j=1}^{n^{[\ell-1]}} m_{ij}^{[\ell]}\; \Var(w_{ij}^{[\ell]})\; \Var(a_{j}^{[\ell-1]})&\\
    & = \Var(w_{ij}^{[\ell]})\; \Var(a_{j}^{[\ell-1]})\, \sum_{j=1}^{n^{[\ell-1]}} m_{ij}^{[\ell]}.&
\end{align}
Let neuron $i$'s number of non-masked weights be denoted $\FanIn_i^{[\ell]}$, where $\FanIn_i^{[\ell]} = \sum_{j=1}^{n^{[\ell-1]}} m_{ij}^{[\ell]}$, then
\begin{align}
    \Var(a_{i}^{[\ell-1]}) & = \FanIn_i^{[\ell]} \Var(w_{ij}^{[\ell]}) \Var(a_{j}^{[\ell-1]})\\
    \textrm{Recall, } \Var(a_{i}^{[\ell-1]}) &= \Var(a_{j}^{[\ell-1]})\nonumber\\
    \Rightarrow \Var(w_{ij}^{[\ell]})& = \frac{1}{\FanIn_i^{[\ell]}}.
\end{align}
Therefore, in order to have the output of each neuron $a_i^{[\ell]}$ in layer $\ell$ to have unit variance, and mean 0, we need to sample the weights for each neuron from the normal distribution,
\begin{equation}
    [w_{ij}^{[\ell]}] \sim  \mathcal{N}\left(0,\,\frac{1}{\FanIn_i^{[\ell]}}\right),
\end{equation}
where $s_i^{\ell}$ is the sparsity of weights of the neuron with output $a_i$. For the ReLU activation function, following the derivation in \citet{He2015delvingdeep},
\begin{equation}
    [w_{ij}^{[\ell]}] \sim  \mathcal{N}\left(0,\,\frac{2}{\FanIn_i^{[\ell]}}\right).
\end{equation}
\subsection{Fixed Mask: Backward Pass}
Given a sparse \gls{nn}, where the output of a neuron $a_{i}$ is given by,
$a_{i}^{[\ell]} = f\left(z_{i}^{[\ell]}\right)$, where $z_{i}^{[\ell]} = \sum_j^{n^{[\ell-1]}} m_{ij}^{[\ell]} w_{ij}^{[\ell]} a_{j}^{[\ell-1]},$
where $m_{ij}^{[\ell]} \in M^{[\ell]}$ and $w_{ij}^{[\ell]} \in W^{[\ell]}$ are the mask and weights respectively for layer $\ell$, and $a_{j}^{[\ell-1]}$ the output of the previous layer. Assume the mask $M^{[\ell]} \in \mathbbm{1}^{n^{[\ell]}\times n^{[\ell - 1]}}$ is constant, where $\mathbbm{1}^{n^{[\ell]}\times n^{[\ell - 1]}}$ is an indicator matrix, and let $L\left(\theta=\{W^{[\ell]}, \ell=0\ldots{}N\}\right)$ be the loss we are optimizing.

As in \citet{Glorot2010understanding}, from the backward-propagation standpoint, we want to ensure $\Var(\frac{\partial L}{\partial z_{i}^{[\ell]}}) = \Var(\frac{\partial L}{\partial z_{i}^{[\ell-1]}}))$, and $\mean(\frac{\partial L}{\partial z_{i}^{[\ell]}}) = 0$. Assume that $f^\prime(0) = 1$,
\begin{align}
    \Var(\frac{\partial L}{\partial z_{j}^{[\ell]}})& \approx \Var(\frac{\partial L}{\partial a_{j}^{[\ell-1]}})&\\
    & = \Var\left(\sum_{i=1}^{n^{[\ell]}} m_{ij}^{[\ell]} w_{ij}^{[\ell]} \frac{\partial L}{\partial z_{i}^{[\ell]}}\right)&\\
    &= \sum_{i=1}^{n^{[\ell]}}\Var\left(m_{ij}^{[\ell]} w_{ij}^{[\ell]} \frac{\partial L}{\partial z_{i}^{[\ell]}}\right)\\
    & \hspace{5em} \textrm{(independent sum)}\nonumber\\
    &= \sum_{i=1}^{n^{[\ell]}} \left(m_{ij}^{[\ell]}\right)^2 \Var\left(w_{ij}^{[\ell]} \frac{\partial L}{\partial z_{i}^{[\ell]}}\right)\\
    & \hspace{3em} \because m_{ij}^{[\ell]} \textrm{ is constant}, \Var(cX) = c^2\,\Var(X)\nonumber\\
    &= \sum_{i=1}^{n^{[\ell]}}\; m_{ij}^{[\ell]}\; \Var\left(w_{ij}^{[\ell]} \frac{\partial L}{\partial z_{i}^{[\ell]}}\right)\\
    & \hspace{5em} \because m_{ij}^{[\ell]} \in [0, 1], \left(m_{ij}^{[\ell]}\right)^2 = m_{ij}^{[\ell]}.\nonumber\\
    & = \sum_{i=1}^{n^{[\ell]}}\; m_{ij}^{[\ell]}\; \Var(w_{ij}^{[\ell]})\; \Var(\frac{\partial L}{\partial z_{i}^{[\ell]}}).\\
    & \hspace{5em} \textrm{(independent product)}\nonumber\label{eqn:backwardbeforeweightassumption}
\end{align}
Assume $\Var(w_{mj}^{[\ell]}) = \Var(w_{nj}^{[\ell]}),\, \forall n,m$, i.e.\ the variance of all weights for a given neuron are the same, and $\Var(\frac{\partial L}{\partial z_{n}^{[\ell]}}) = \Var(\frac{\partial L}{\partial z_{m}^{[\ell]}})$, i.e.\ the variance of the output gradients of each neuron at layer $l$ are the same. Then we can simplify \cref{eqn:backwardbeforeweightassumption}, 
\begin{align}
    \Var(\frac{\partial L}{\partial z_{j}^{[\ell]}})& = \sum_{i=1}^{n^{[\ell]}} m_{ij}^{[\ell]}\; \Var(w_{ij}^{[\ell]})\; \Var(\frac{\partial L}{\partial z_{i}^{[\ell]}})&\\
    & = \Var(w_{ij}^{[\ell]})\; \Var(\frac{\partial L}{\partial z_{i}^{[\ell]}})\, \sum_{i=1}^{n^{[\ell]}} m_{ij}^{[\ell]}.&
\end{align}
Let neuron $i$'s number of non-masked weights be denoted $\FanOut_j^{[\ell]}$, where $\FanOut_j^{[\ell]} = \sum_{i=1}^{n^{[\ell]}} m_{ij}^{[\ell]}$, then
\begin{align}
    \Var(\frac{\partial L}{\partial z_{j}^{[\ell]}}) & = \FanOut_j^{[\ell]} \Var(w_{ij}^{[\ell]}) \Var(\frac{\partial L}{\partial z_{i}^{[\ell]}})\\
    \textrm{Recall, } \Var(\frac{\partial L}{\partial z_{j}^{[\ell]}}) &= \Var(\frac{\partial L}{\partial z_{i}^{[\ell]}})\nonumber\\
    \Rightarrow \Var(w_{ij}^{[\ell]})& = \frac{1}{\FanOut_j^{[\ell]}}.
\end{align}
Therefore, in order to have the output of each neuron $a_i^{[\ell]}$ in layer $\ell$ to have unit variance, and mean 0, we need to sample the weights for each neuron from the normal distribution,
\begin{equation}
    [w_{ij}^{[\ell]}] \sim  \mathcal{N}\left(0,\,\frac{1}{\FanIn_i^{[\ell]}}\right),
\end{equation}
where $s_i^{\ell}$ is the sparsity of weights of the neuron with output $a_i$. For the ReLU activation function, following the derivation in \citet{He2015delvingdeep},
\begin{equation}
    [w_{ij}^{[\ell]}] \sim  \mathcal{N}\left(0,\,\frac{2}{\FanIn_i^{[\ell]}}\right).
\end{equation}
\section{Additional Gradient Flow Plots}
\label{sec:gradnorm_app}
Here we share additional gradient flow figures for method/initialization combinations presented in \ref{fig:dynamicsparse} and \ref{fig:gradnorm}.

\begin{figure*}[tp]
  \centering
  \begin{subfigure}{.33\textwidth}
  \centering
  \includegraphics[width=\linewidth]{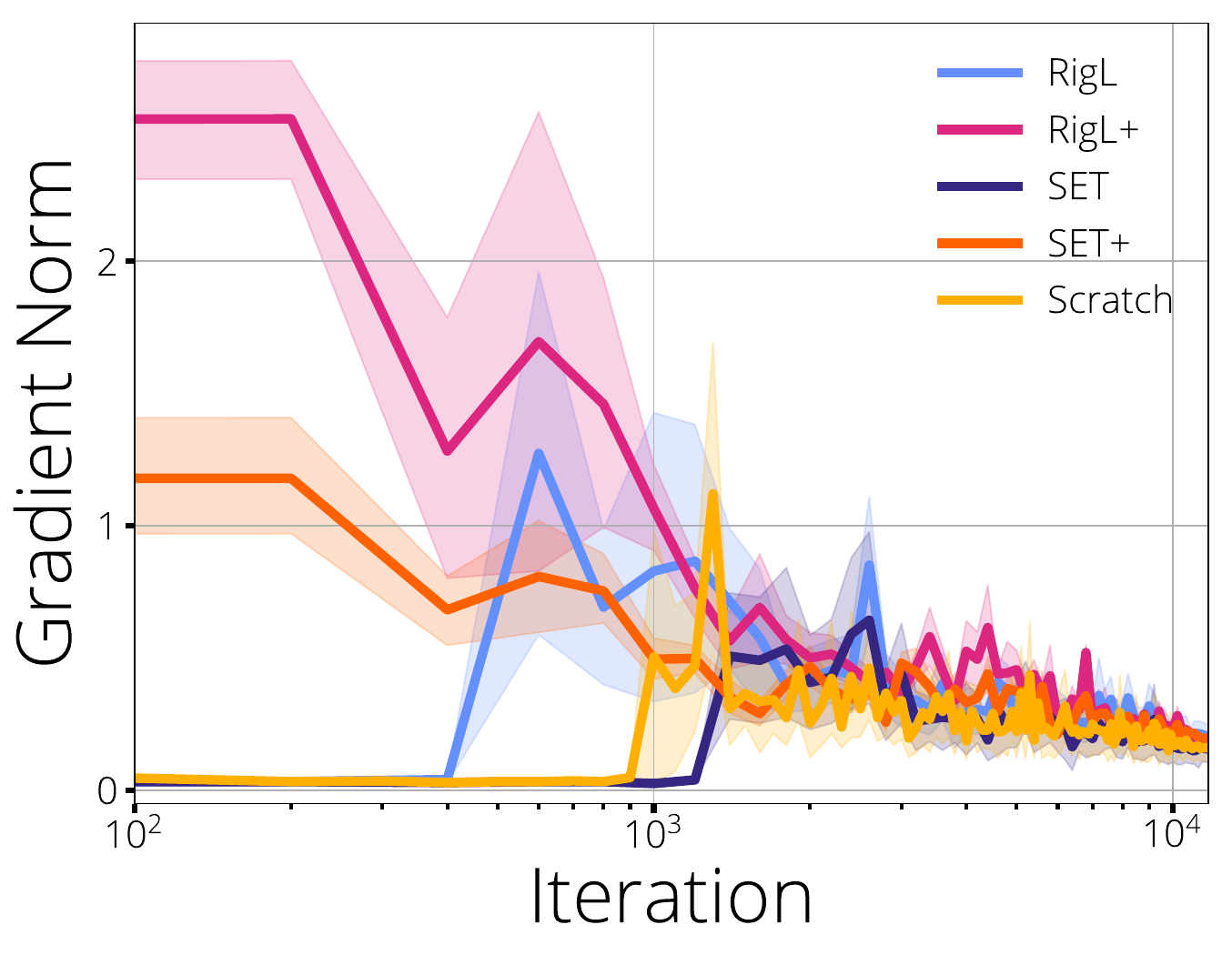}
      \caption{}%
\label{subfig:gradnorm1_app}
\end{subfigure}%
    \begin{subfigure}{.33\textwidth}
  \centering
\includegraphics[width=\linewidth]{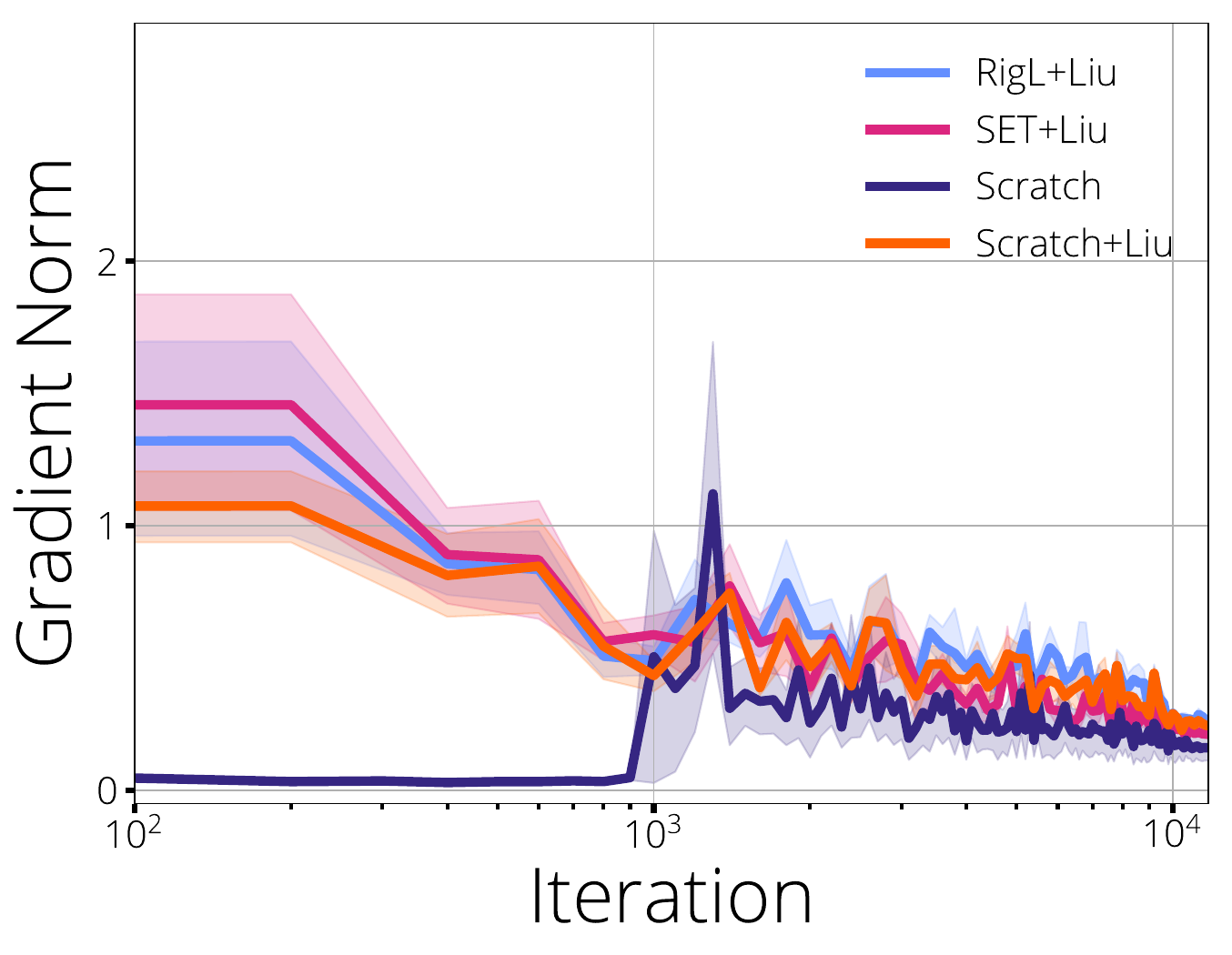}
    \caption{}%
\label{subfig:gradnorm2_app}
\end{subfigure}%
\begin{subfigure}{.33\textwidth}
  \centering
\includegraphics[width=\linewidth]{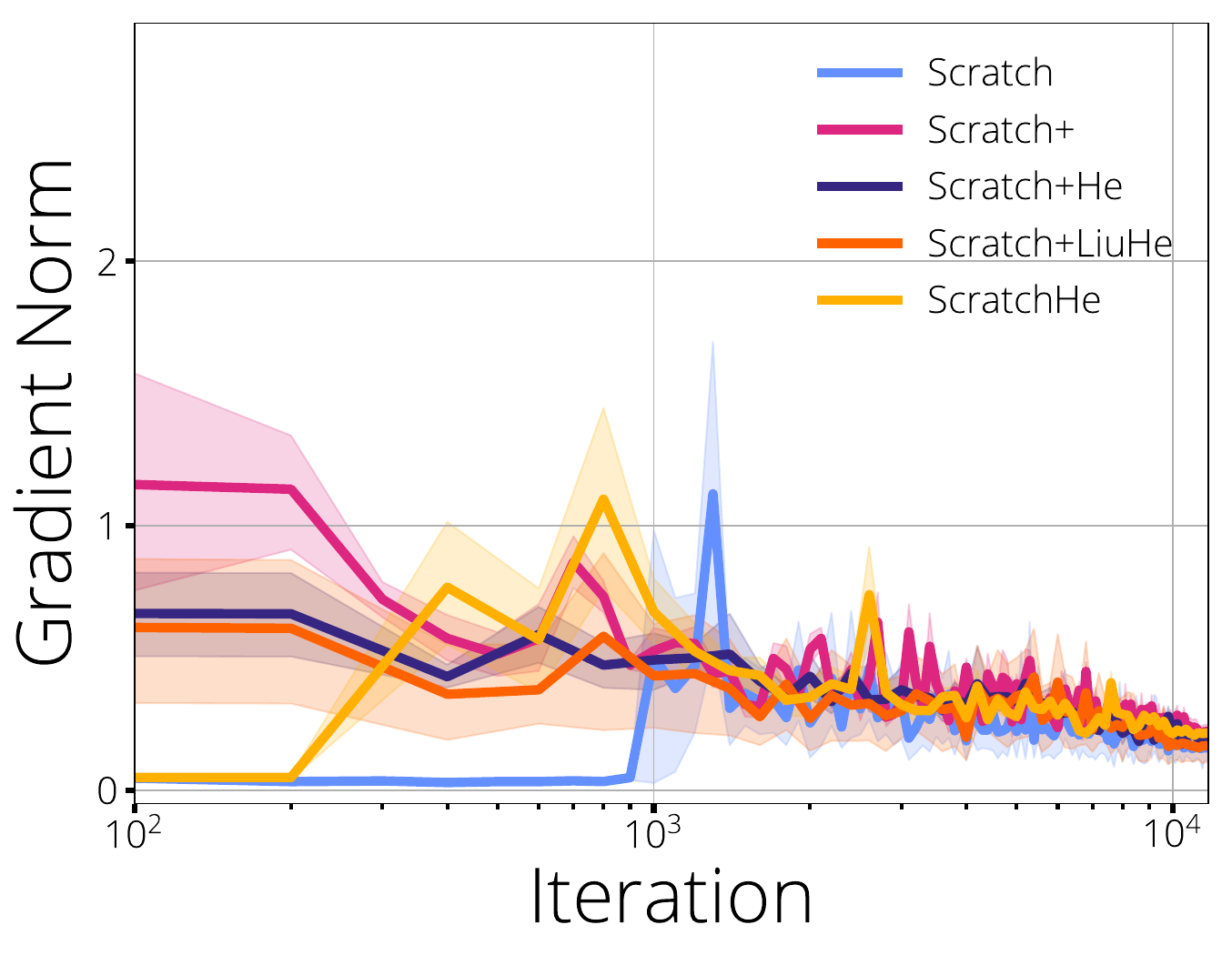}
    \caption{}%
\label{subfig:gradnorm3_app}
\end{subfigure}%
    \caption{\textbf{Gradient Flow of Sparse LeNet-5s during Training.} Gradient flow during training averaged over multiple runs, `+` indicates training runs with our proposed sparse initialization. `+Liu` indicates initialization proposed by \cite{Liu2018}. `He` suffix refers to He initilization where `scale=2` and `fanin` options are used for the variance scaling initialization.}
    \label{fig:gradnorm_app}
\end{figure*}

In \cref{subfig:gradnorm2_app}, we show the gradient flow for \gls{dst} methods and scratch training. Using \gls{rigl} helps improves gradient flow with both initialization; helping learning to start earlier than regular \textit{Scratch} training. \gls{set} seem to have limited effect on the gradient flow.  

In \cref{subfig:gradnorm2_app}, we share gradient flow when the scaled initialization of \cite{Liu2018} is used. Similar to the proposed initialization, we observe improved gradient flow for all cases. Different than our initialization however, \gls{rigl} doesn't improve gradient flow in this setting; highlighting an interesting future research direction on the relationship between initialization and the \gls{dst} methods.

In \cref{subfig:gradnorm3_app}, we share gradient flow when He initialization is used instead of Glorot initialization. We observe that \textit{Scratch} training starts learning faster in this case. Gradient flow seems to be similar for other sparse initialization methods.

\begin{figure*}[tp]
\centering
\begin{subfigure}[t]{.45\linewidth}
\includegraphics[width=\linewidth]{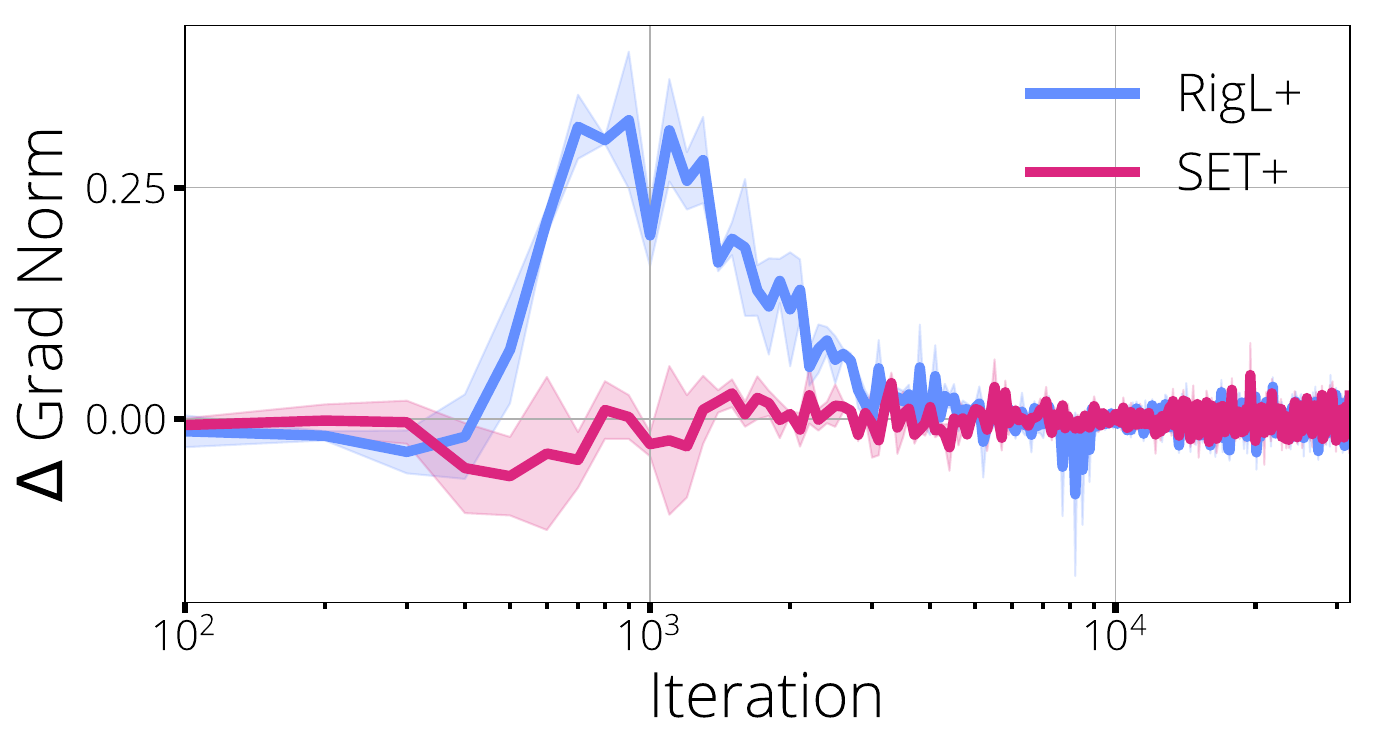}
\caption{ResNet-50}%
\label{subfig:dynamicsparse1_app}
\end{subfigure}%
\begin{subfigure}[t]{.45\linewidth}
\includegraphics[width=\linewidth]{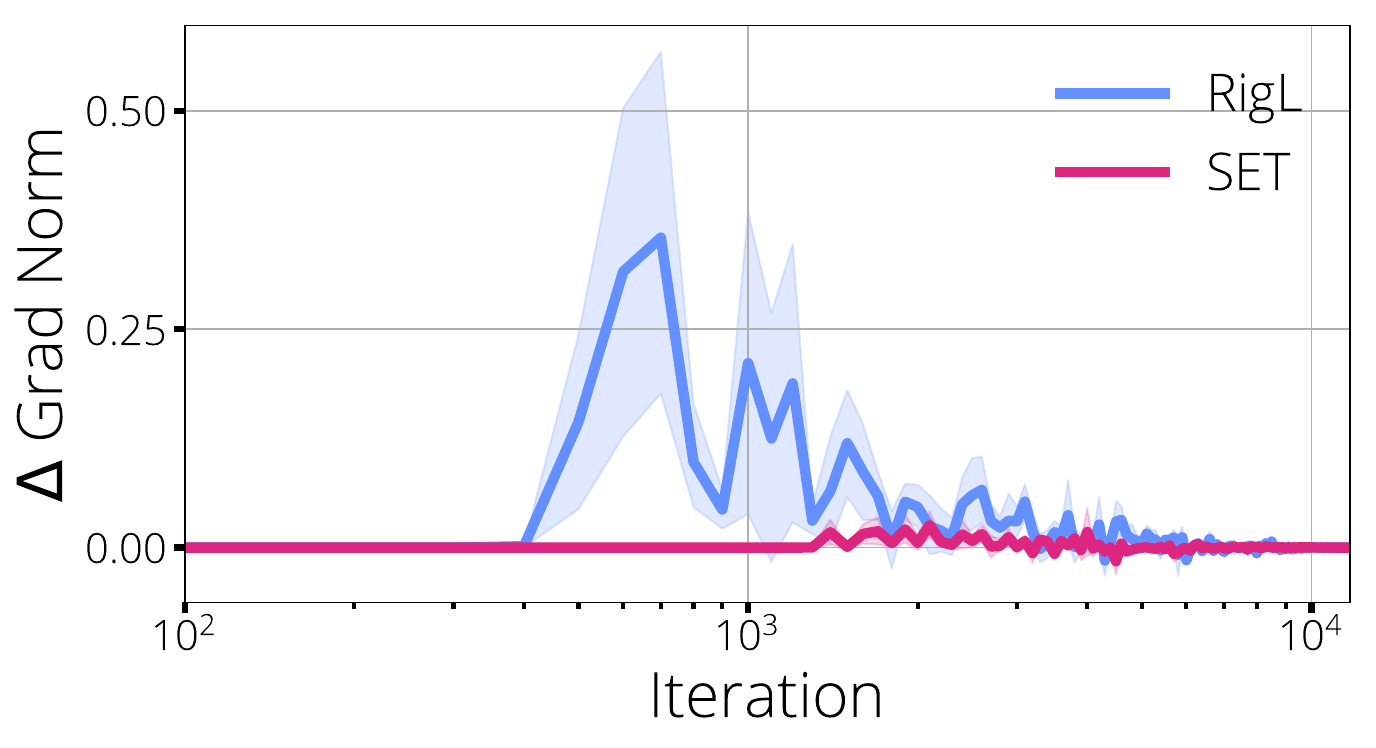}
\caption{LeNet5}%
\label{subfig:dynamicsparse2_app}
\end{subfigure}%

\caption{\textbf{Effect of Mask Updates in \acrlong{dst}.} Effect of mask updates on the gradient norm. We measure the gradient norm before and after the mask updates and plot the $\Delta$. `+` indicates proposed initialization and used in MNIST experiments.}\label{fig:dynamicsparse_app}
\end{figure*}

In \cref{fig:dynamicsparse_app}, we share gradient flow improvements after \gls{dst} updates on connectivity for different initialization methods. ResNet-50 curves match the results in \cref{fig:dynamicsparse}. LeNet5 curves however seem to be adversely affected by poor initialization at the beginning of the training. We start observing improvements with \gls{rigl} when the learning starts (around the \engordnumber{400} iteration).


\section{Fully Connected Neural Network Experiments on MNIST}
\label{sec:mlp_mnist_app}
In this section we repeat our experiments from \cref{sec:exp_init} using different sparse initialization methods, and analyzing gradient flow, for a standard 2-layer fully-connected \gls{nn} with 2 hidden layers of size 300 and 100 units. We use the same grid used in LeNet5 experiments for hyper-parameter selection. Best results were obtained with a learning rate of 0.2, a weight decay coefficient of 0.0001 and an mask update frequency of 500 (used in \gls{dst} methods). The rest of the hyperparameters remained unchanged from the LeNet5 experiments.

\begin{table}[tp]
\centering
\caption{\textbf{Results of Trained Fully-Connected MNIST Model from Different Initializations.}}%
\label{table:mnistfcinitresults}
\begin{tabular}{@{}lrrr@{}}
\toprule
& \multicolumn{3}{c}{MNIST} \\
\cmidrule(lr){2-4} 
& \multicolumn{3}{c}{Fully-Connected \gls{nn} (98\% sparse)}\\
\cmidrule(lr){2-4}
Baseline & \multicolumn{3}{c}{98.55$\pm$0.04}\\
Lottery & \multicolumn{3}{c}{97.73$\pm$0.11}\\
Small Dense & \multicolumn{3}{c}{91.69$\pm$1.85}\\[1.2ex]
& \multicolumn{1}{c}{Original} & \multicolumn{1}{c}{Liu et al.} & \multicolumn{1}{c}{Ours}\\
\cmidrule(lr){2-2} \cmidrule(lr){3-3} \cmidrule(lr){4-4}
Scratch & 94.40$\pm$4.00 & 96.66$\pm$0.18 & \textbf{96.70}$\pm$0.12\\
\acrshort{set} & 96.49$\pm$0.36 &  \textbf{96.56}$\pm$0.22 & 96.48$\pm$0.10\\
\acrshort{rigl} & 96.82$\pm$0.25 & 96.76$\pm$0.19 & \textbf{96.94}$\pm$0.12\\ 
\bottomrule
\end{tabular}
\end{table}
The results of training with various initialization methods is shown in \cref{table:mnistfcinitresults}. Although the results are not as drastic as with LeNet5, we see that here too sparsity aware initialization (the proposed initialization, and that of Liu~\citep{Liu2018}) shows a significant improvement in the test accuracy of Scratch, and \gls{rigl} or our proposed initialization, although not quite reaching lottery or \gls{rigl} accuracy. Finally, we see no significant effect on \gls{set} training, with none of the initialization variants having a significant increase over any of the others, although the Liu~\citep{Liu2018} initialization does marginally better.

\begin{figure*}[tp]
  \centering
  \begin{subfigure}{.33\textwidth}
  \centering
  \includegraphics[width=\linewidth]{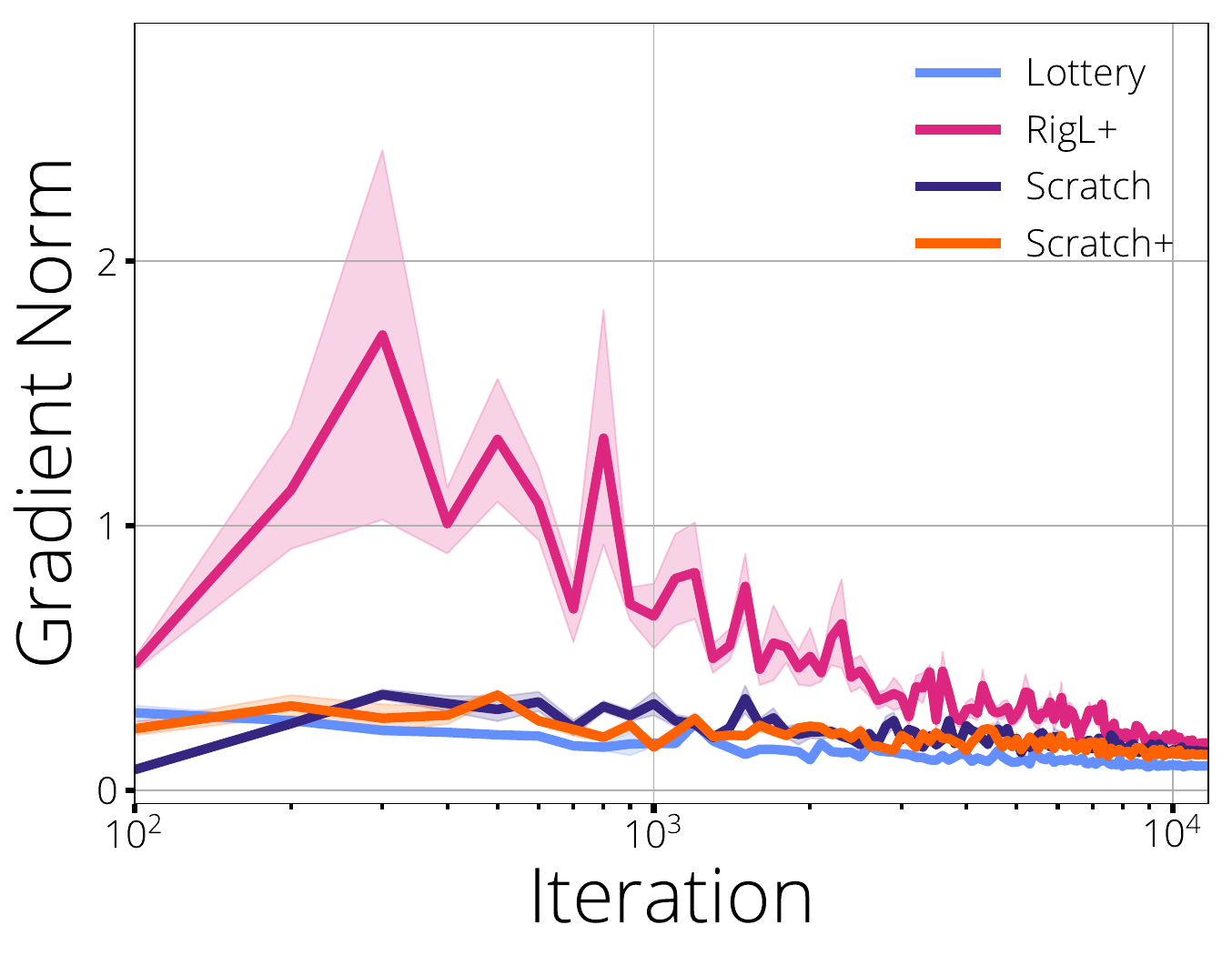}
      \caption{Gradient Flow during training}%
\label{subfig:mnist_mlp_app}
\end{subfigure}%
    \begin{subfigure}{.33\textwidth}
  \centering
\includegraphics[width=\linewidth]{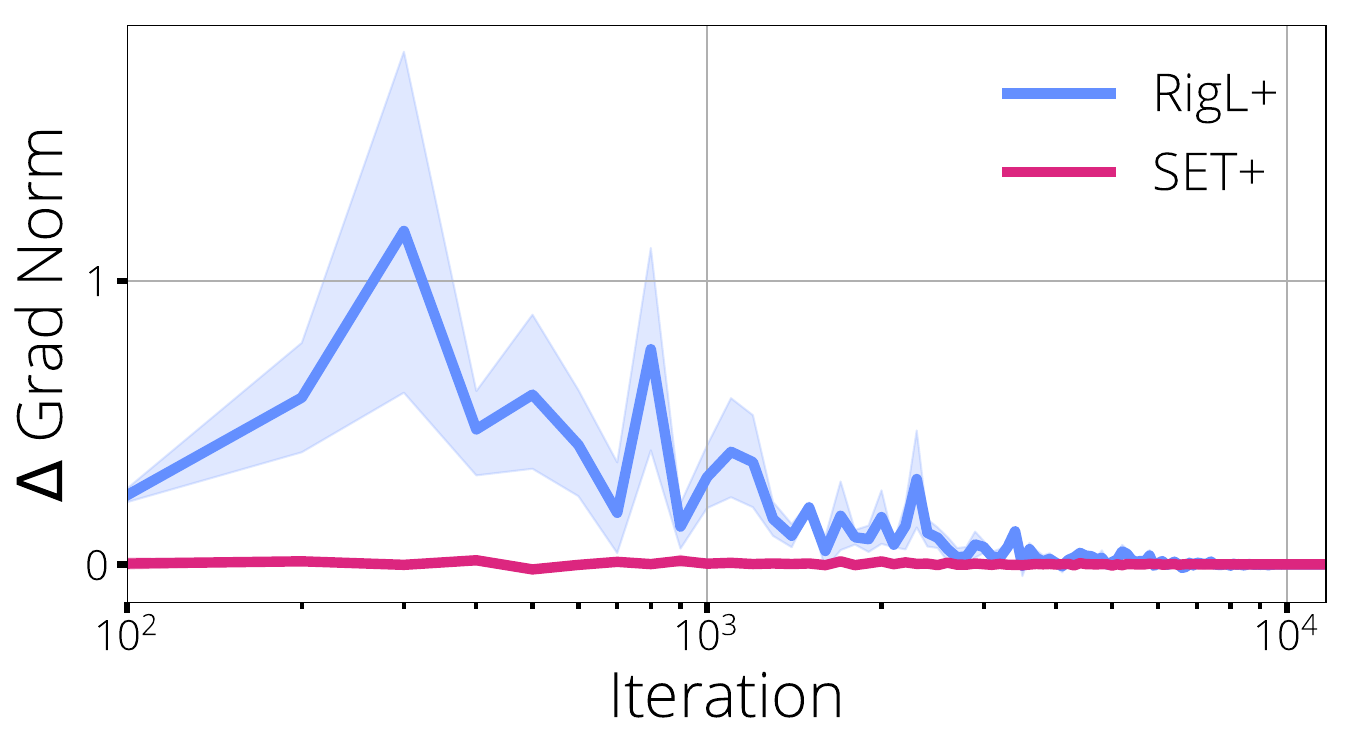}
    \caption{Gradient Flow improvements after connectivity updates}%
\label{subfig:mnist_mlp_app2}
\end{subfigure}%
\begin{subfigure}{.33\textwidth}
  \centering
\includegraphics[width=\linewidth]{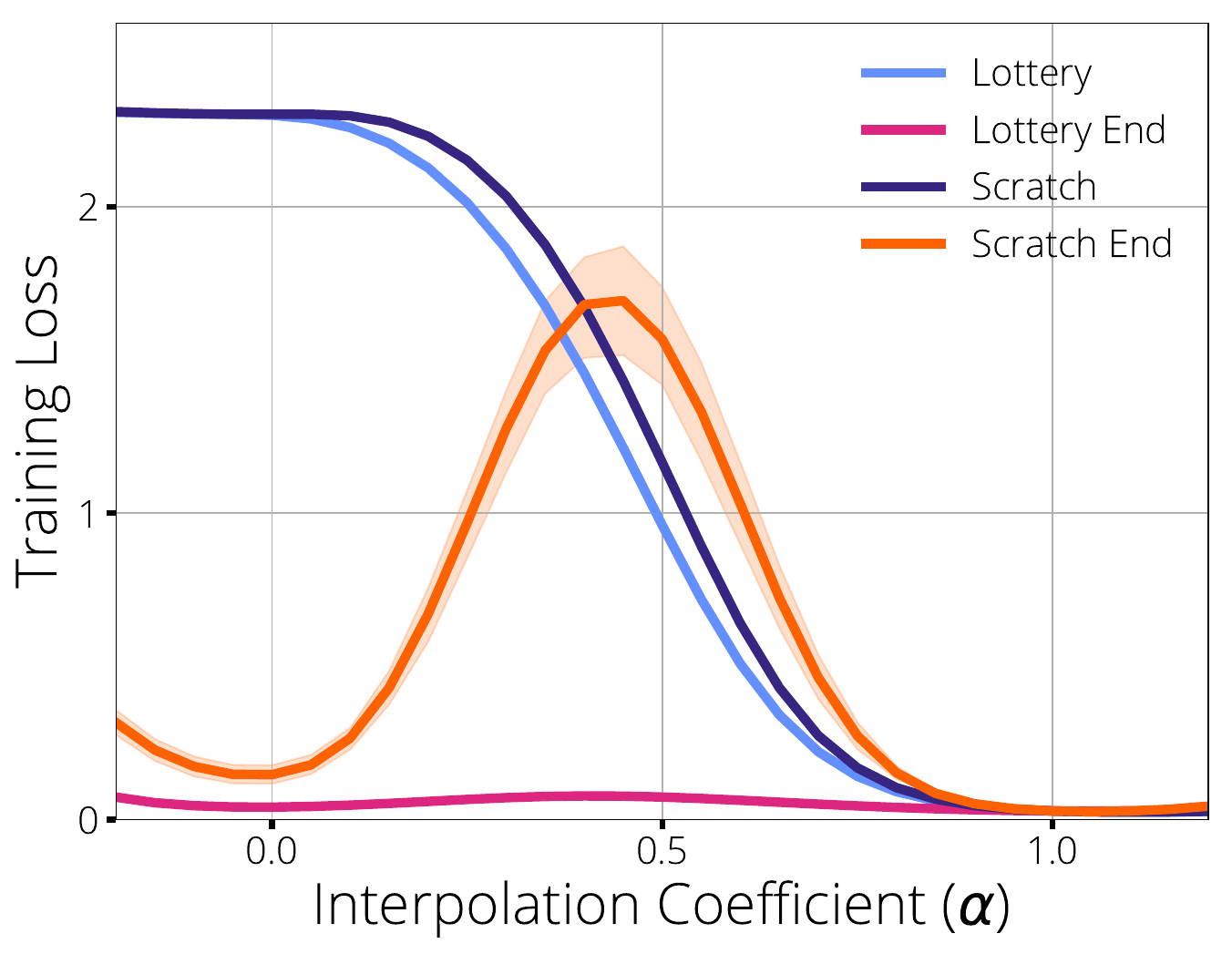}
    \caption{Interpolation}%
\label{subfig:mnist_mlp_app3}
\end{subfigure}%
    \caption{\textbf{Sparse 300-100 MLP experiments.} Gradient flow during training averaged over multiple runs, `+` indicates training runs with our proposed sparse initialization.}
    \label{fig:mnist_mlp_app}
\end{figure*}

The gradient flow of this model is shown in \cref{fig:mnist_mlp_app}. While we see moderate improvements to Scratch gradient flow early on in training with our proposed initialization (\subref{subfig:mnist_mlp_app}), \gls{rigl} shows significantly higher gradient flow throughout training, in particular after mask updates (\subref{subfig:mnist_mlp_app2}), mirroring the results of LeNet5. The interpolation graphs in (\subref{subfig:mnist_mlp_app3}) only differ slightly from that of LeNet5, again showing that our results for LeNet5 broadly hold for the fully-connected model.

\section{Hessian Spectrum of LeNet5}
\label{sec:lenet5hessian}
Given a loss function $L$ and parameters $\theta$, we can write the first order Taylor approximation of the change in loss $\Delta L=L(\theta^{t+1})-L(\theta^{t})$ after a single training step with the learning rate $\epsilon>0$ as :
\begin{equation}
\Delta L \approx -\epsilon \nabla L(\theta)^T\nabla L(\theta).
\end{equation}
Note that as long as the error is small, gradient descent is guaranteed to decrease the loss by an amount proportional to $\nabla L(\theta)^T\nabla L(\theta)$, which we refer as the \emph{gradient flow}. In practice large learning rates are used, and the first order approximation might not be accurate. Instead we can look at the second order approximation of $\Delta L$:
\begin{equation}
\Delta L \approx -\alpha \nabla L(\theta)^T\nabla L(\theta) + \frac{\alpha^2}{2}\nabla L(\theta)^TH(\theta)\nabla L(\theta),\label{eq:secondorderapprox}
\end{equation}
where $H(\theta)$ is the Hessian of the loss function. The eigenvalue spectrum of Hessian can help us understand the local landscape~\citep{sagun2017}, and help us identify optimization difficulties~\citep{ghorbani2019}. For example, if and when the gradient is aligned with large magnitude eigenvalues, the second term of \cref{eq:secondorderapprox} can have a significant effect on the optimization of $L$. If the gradient is aligned with large positive eigenvalues, it can prevent gradient descent from decreasing the loss and harm the optimization. Similarly, if it is aligned with negative eigenvalues it can help to accelerate optimization.

\begin{figure*}[tp]
  \centering
\includegraphics[width=.48\linewidth]{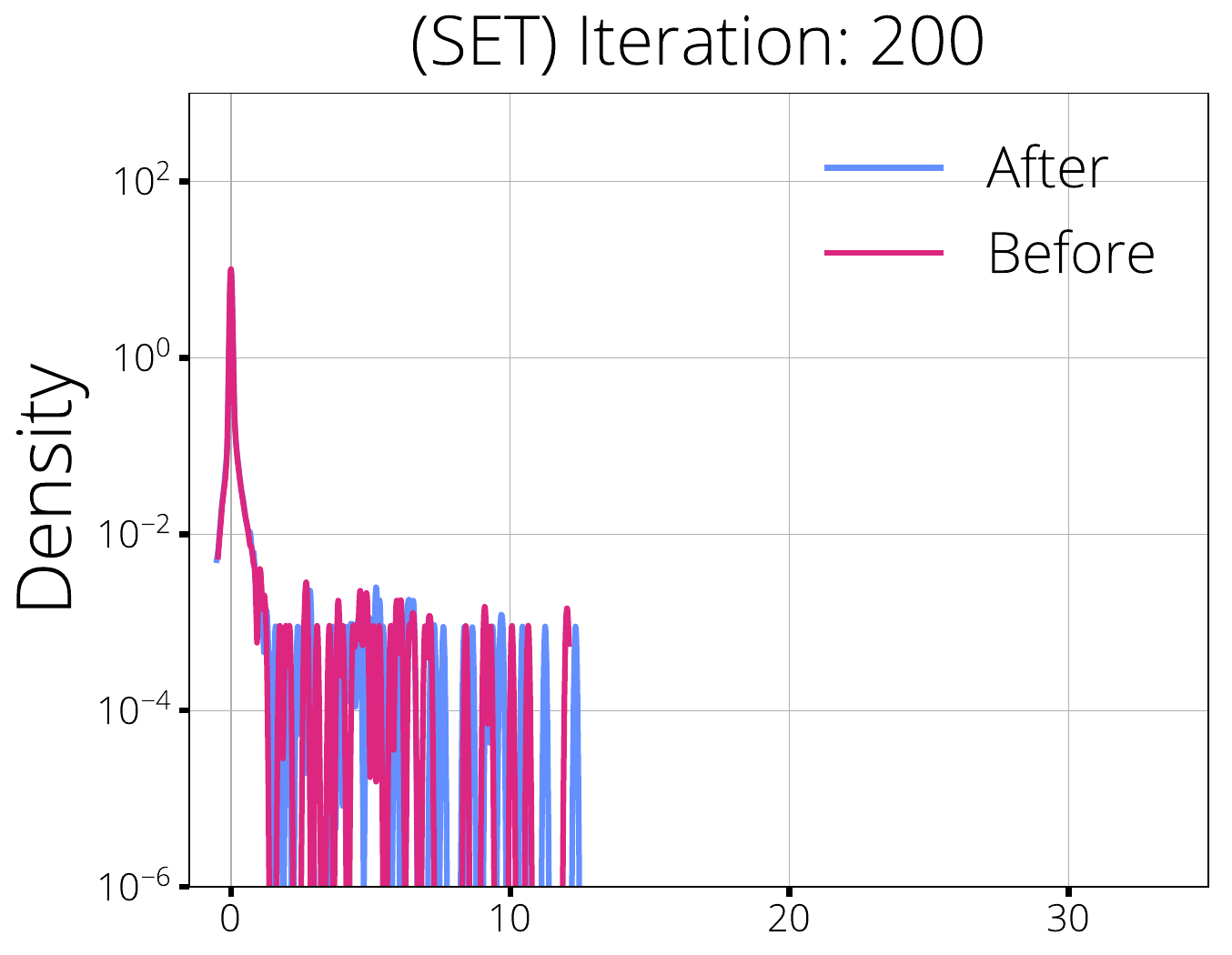}
\includegraphics[width=.48\linewidth]{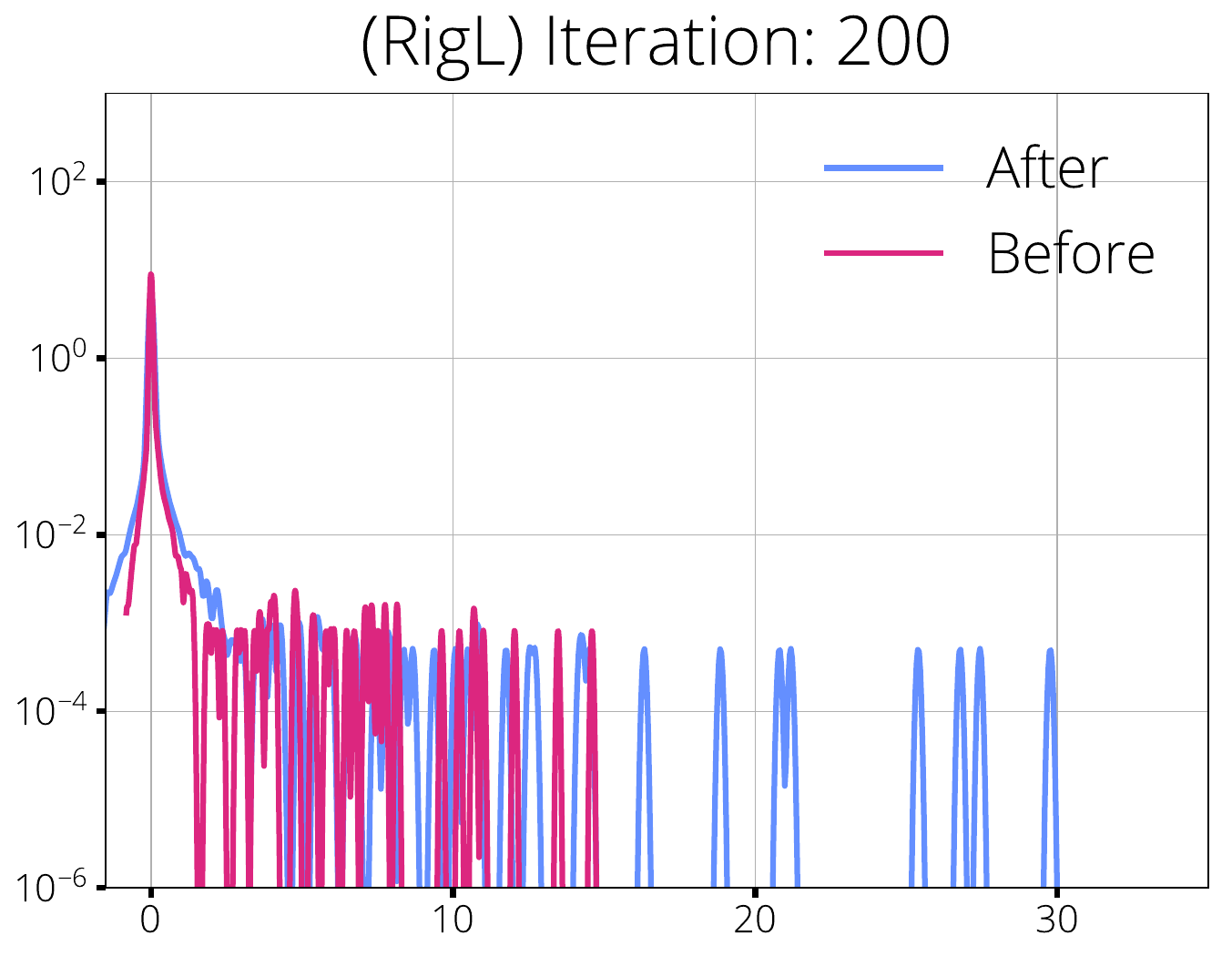}
\includegraphics[width=.48\linewidth]{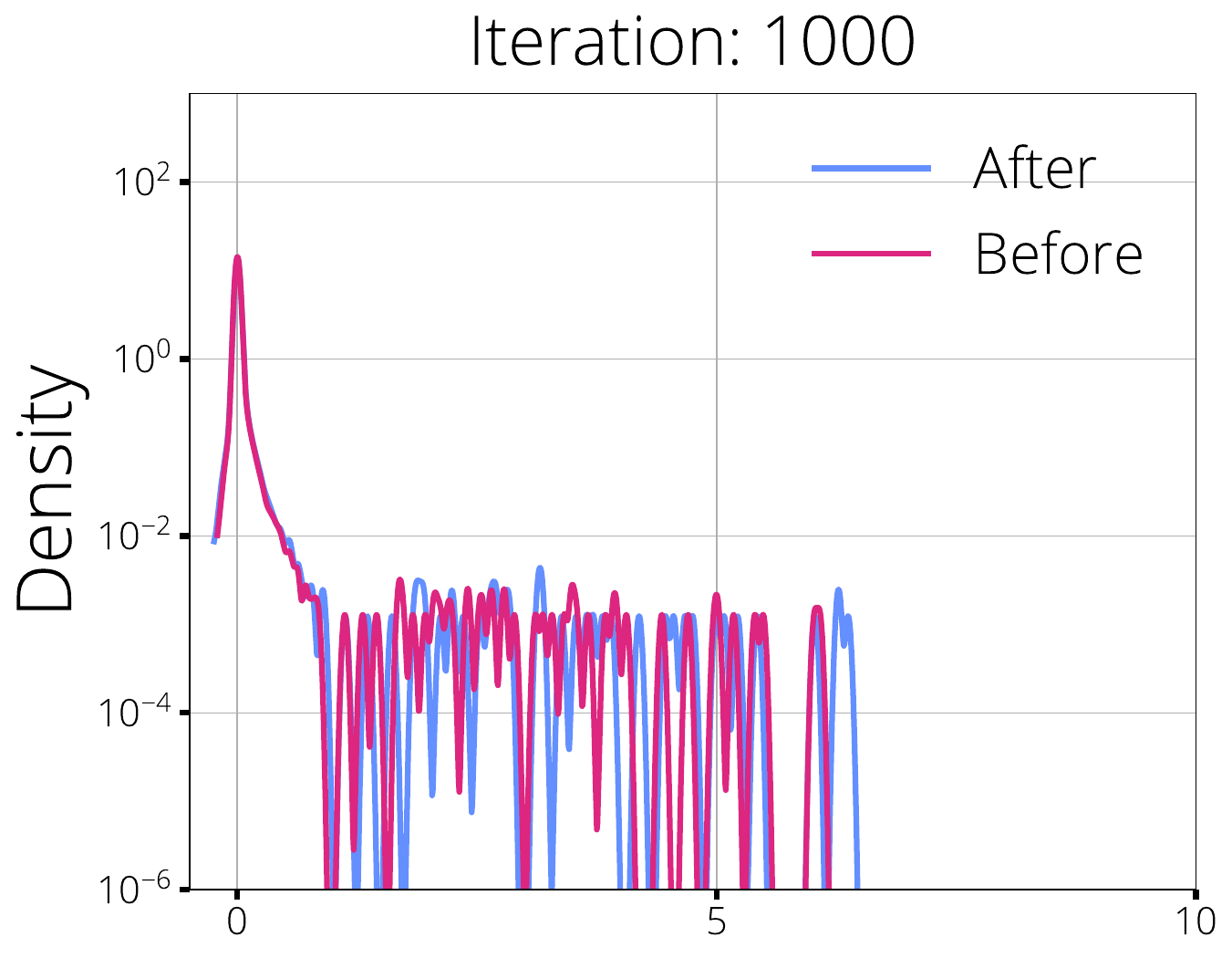}
\includegraphics[width=.48\linewidth]{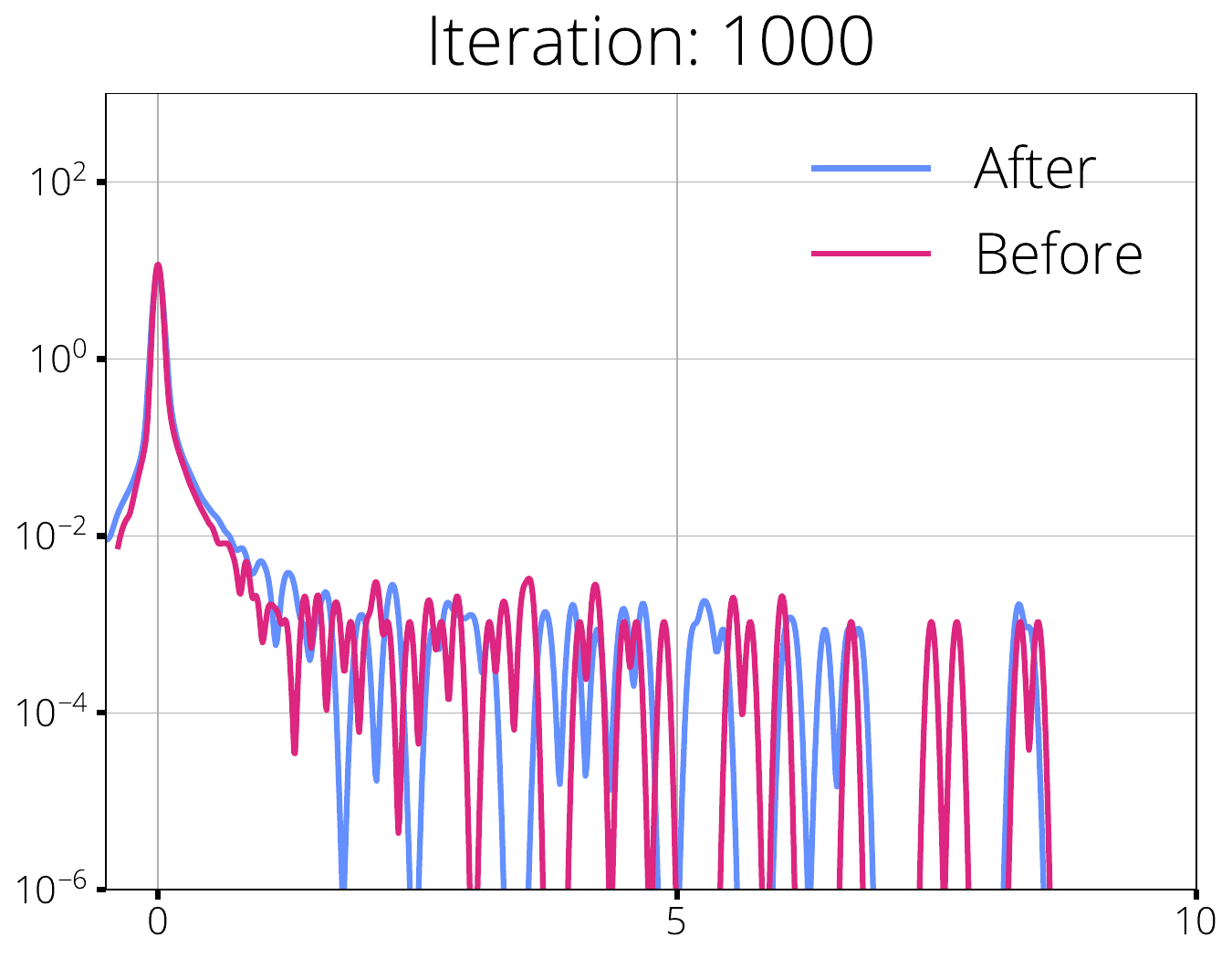}
  \caption{Hessian spectrum before and after mask updates: \textbf{(left)}  SET \textbf{(right)} RigL. Similar to \cite{ghorbani2019}, we estimate the spectral density of Hessian using Gaussian kernels. }%
  \label{fig:hessian_mask_updates}
\end{figure*}

We show the Hessian spectrum before and after the topology updates in \cref{fig:hessian_mask_updates}. After \gls{rigl} updates we observe new negative eigenvalues with significantly larger magnitudes. We also see larger positive eigenvalues, which disappear after few iterations\footnote{We share videos of these transitions in supplementary material.}. In comparison, the effect of \gls{set} updates on the Hessian spectrum seems limited. 
\begin{figure*}[tp]
  \centering
    \begin{subfigure}{.95\textwidth}
  \centering
  \includegraphics[width=.5\linewidth]{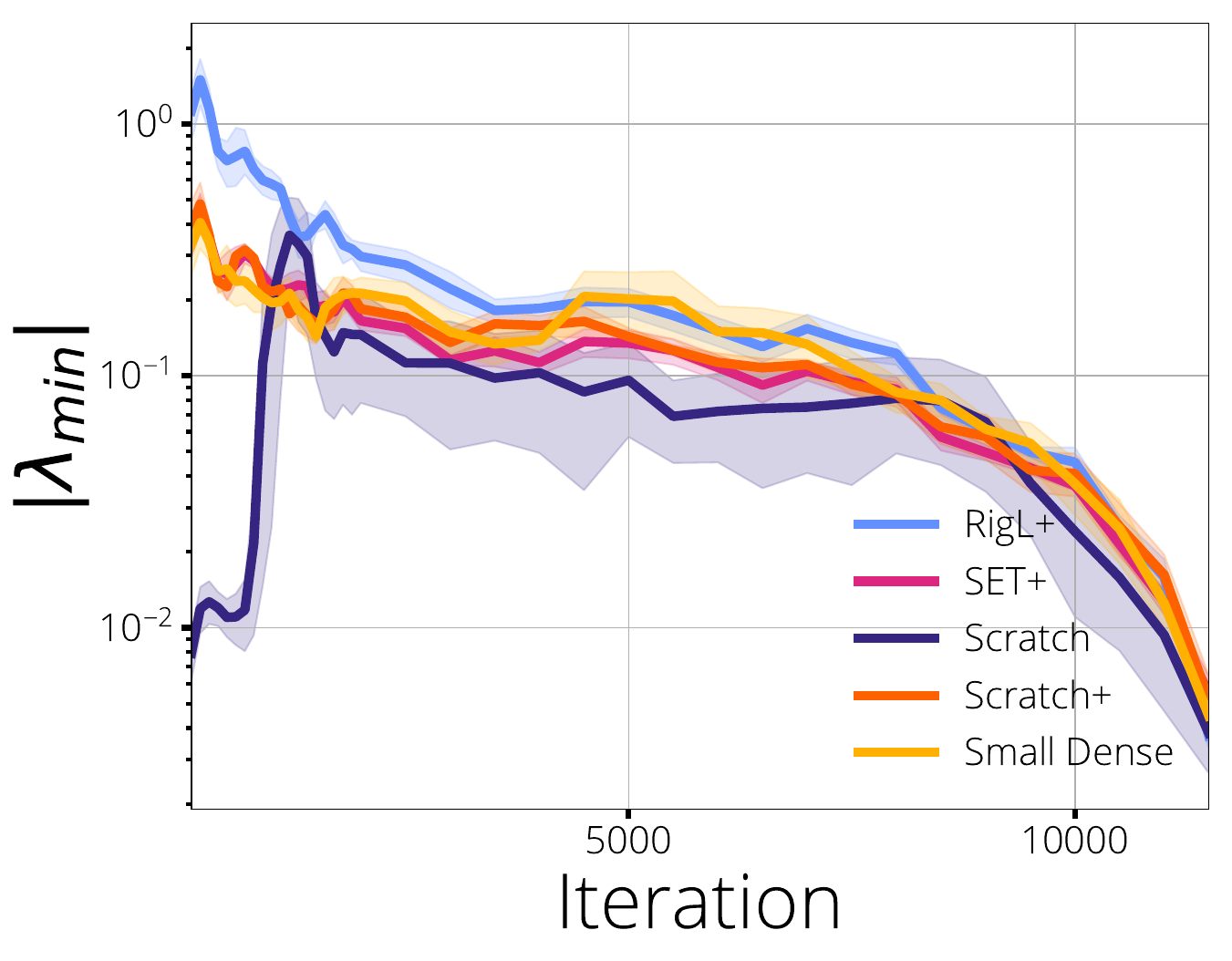}
  \caption{}%
\label{subfig:hessian1}
\end{subfigure}\\
\begin{subfigure}{.95\textwidth}
  \centering
  \includegraphics[width=.45\linewidth]{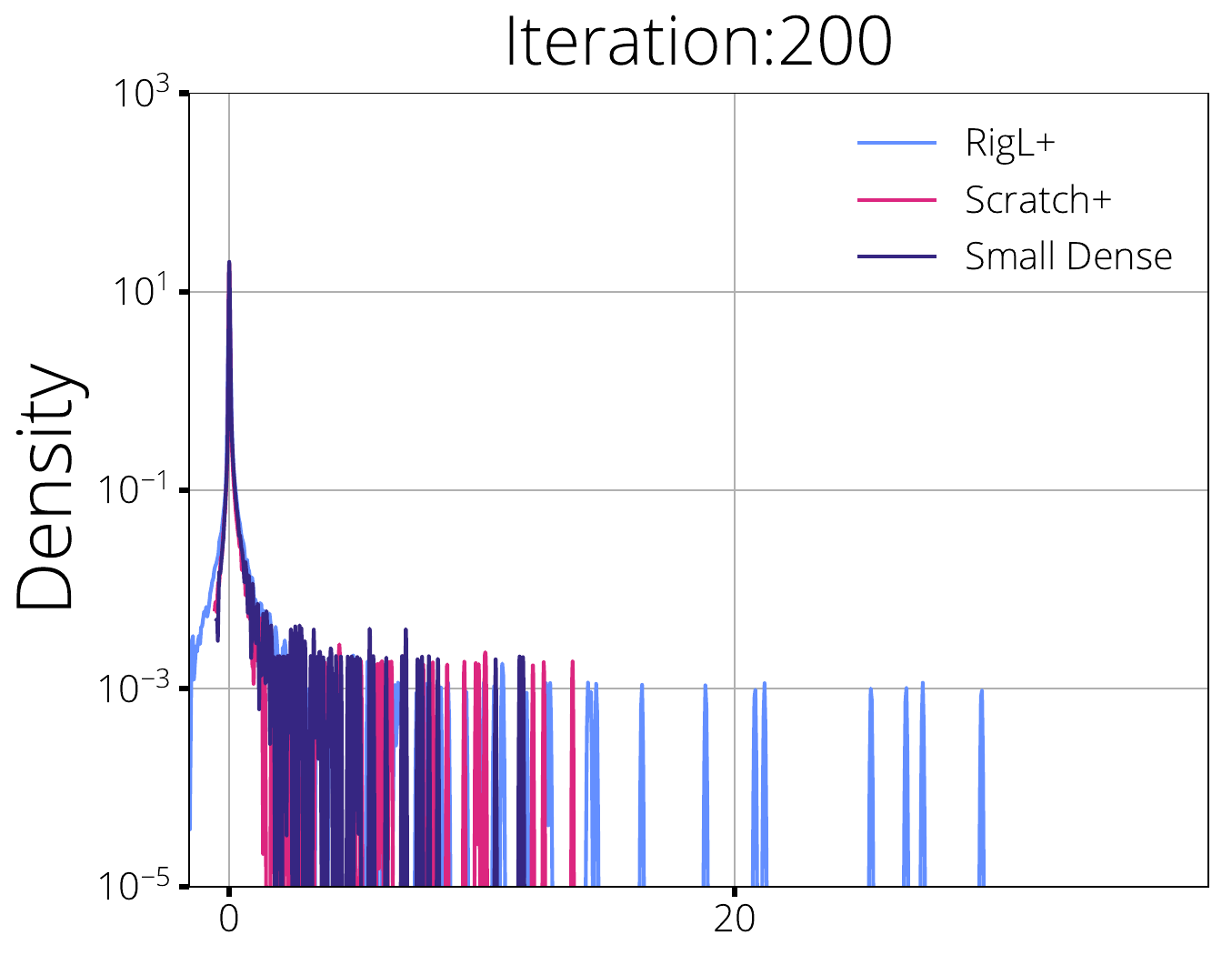}
  \includegraphics[width=.45\linewidth]{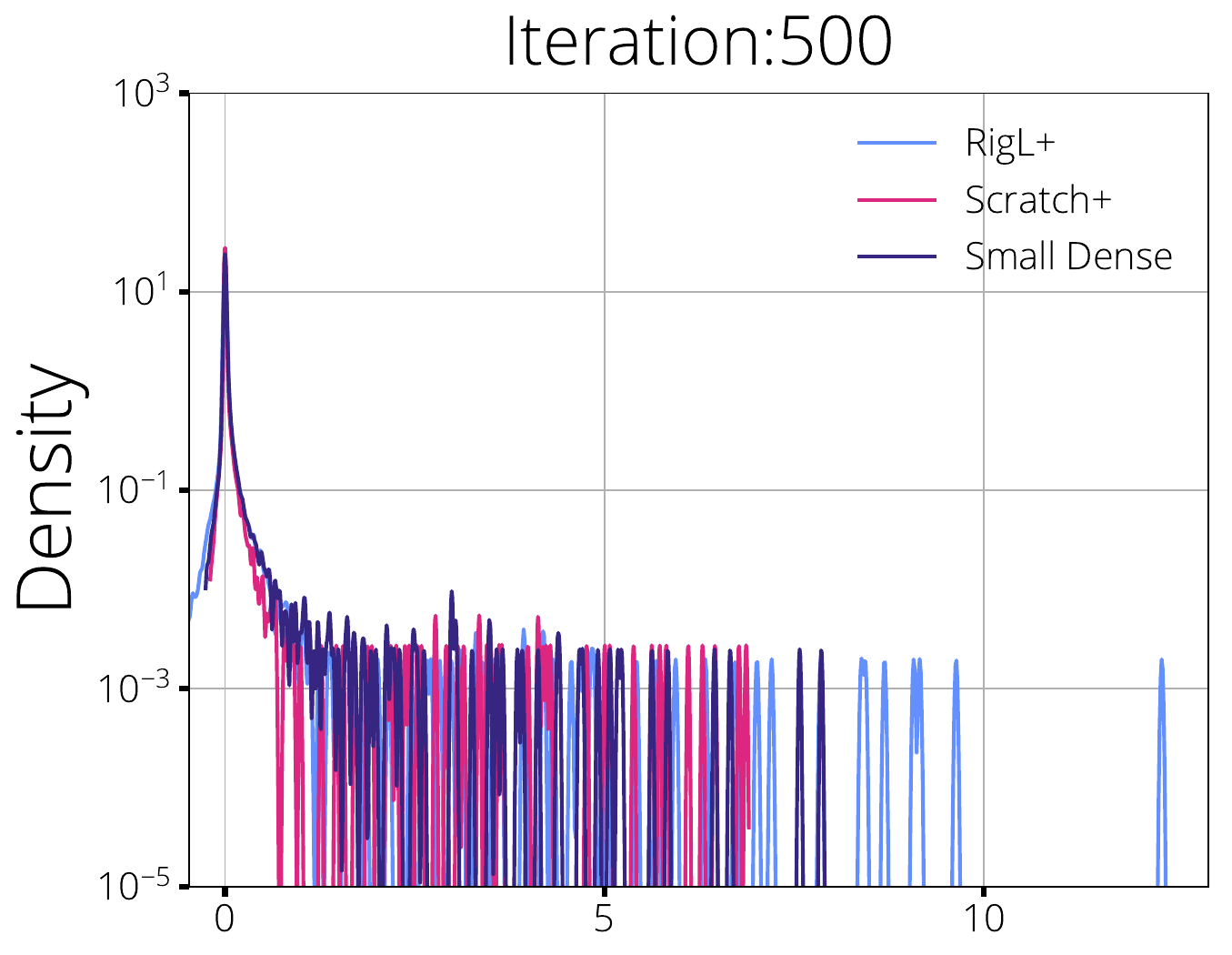}
  \includegraphics[width=.45\linewidth]{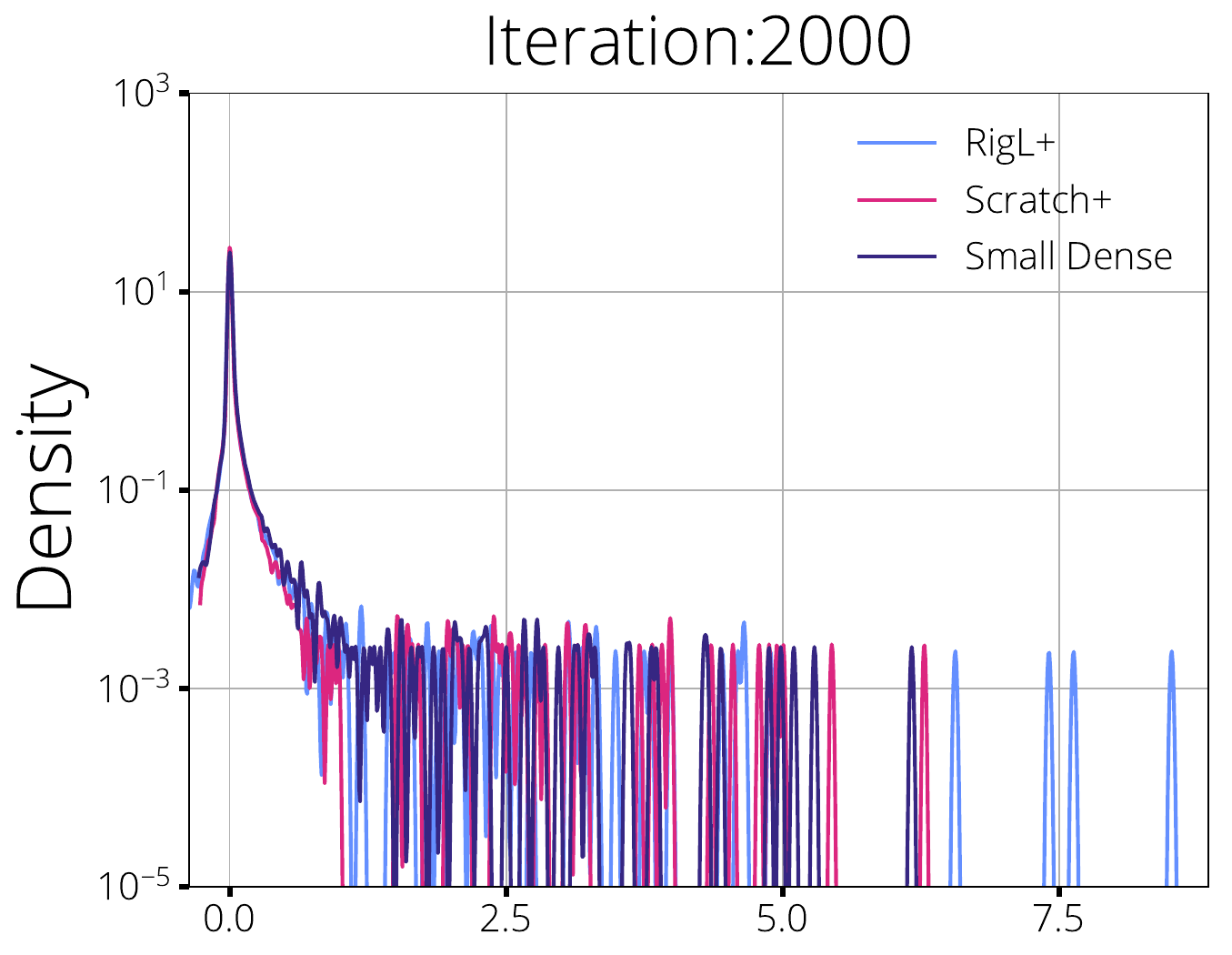}
  \includegraphics[width=.45\linewidth]{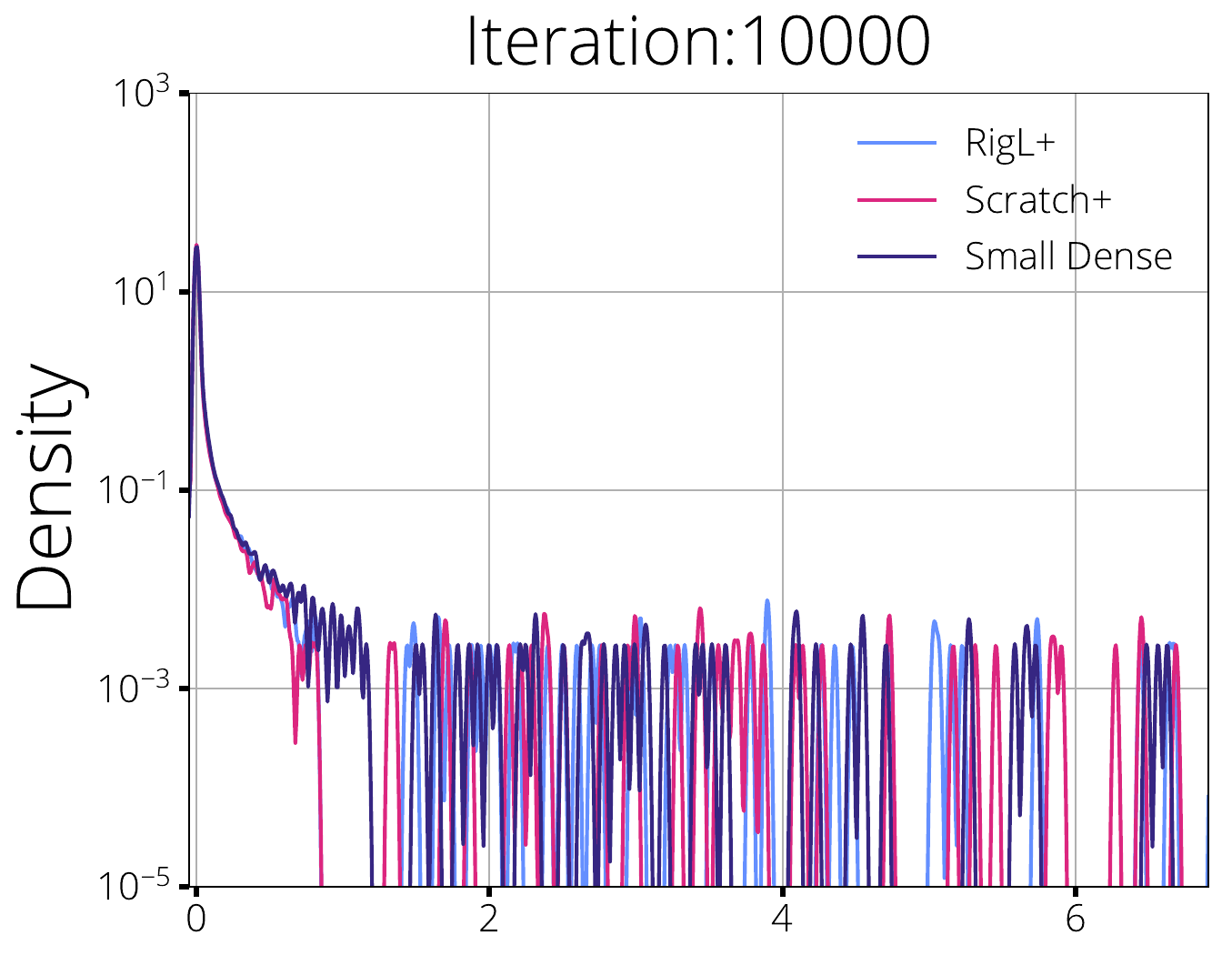}
    \caption{}%
\label{subfig:hessian2}
\end{subfigure}%
  \caption{MNIST Hessian spectrum experiments.}%
  \label{fig:hessian}
\end{figure*}

We also evaluate the Hessian spectrum of LeNet5 during the training. In \cref{subfig:hessian2}, we observe similar shapes for each method on the positive side of the spectrum, however, on the negative side dense models seem to have more mass. We plot the magnitude of the largest negative eigenvalue to characterize this behaviour in \cref{subfig:hessian1}. We observe a significant difference between sparse and dense models and observe that sparse networks trained with \gls{rigl} have larger negative eigenvalues.
\section{Comparing Function Similarity}\label{sec:disagreementfull}
\begin{sidewaystable*}[p]

\footnotesize
\begin{threeparttable}
\caption{\textbf{Ensemble/Prediction Disagreement}. In order to show the function similarity of \glspl{lt} to the pruning solution, we follow the analysis of \citep{Fort2020deep}, and compare the function similarity and ensemble generalization over 5 sparse models trained using random initializations and \glspl{lt} with the original pruning solution they are derived from. The fractional disagreement is the pairwise disagreement of class predictions over the test set, as compared within the group of sparse models, and as compared to the pruned model whose mask they were derived from. \Gls{kl} and \gls{jsd} compare the prediction distributions over all the test samples.}\label{disagreementfull}
\begin{tabular}{@{}lS[table-format=2.3(4)]S[table-format=2.2]S[table-format=1.4(4)]S[table-format=1.4(4)]S[table-format=1.4(4)]S[table-format=1.4(4)]S[table-format=1.5(4)]S[table-format=1.4(4)]@{}}
\toprule
\multicolumn{3}{l}{MNIST/LeNet5}&&&&&&\\
\midrule
\textbf{Init.\ Method} & \textbf{Test Acc. (Top-1)} & \textbf{Ensemble} & \textbf{Pairwise Disagree.} & \textbf{Disagree. w/ Pruned} & \textbf{Pairwise \acrshort{kl}} & \textbf{\acrshort{kl} w/ Pruned} & \textbf{5-model \acrshort{jsd}} & \textbf{\acrshort{jsd} w/ Pruned}\\
\midrule
\textbf{Pruned Soln.}\tnote{*} & 98.53 & {--} & {--} & {--} & {--} & {--} & {--} & {--}\\
\textbf{\gls{lt}} & 98.52 \pm 0.02 & 98.58 & 0.0043\pm 0.0006 & 0.0089 \pm 0.0002 & 0.0030\pm 0.0004 & 0.0039 \pm 0.0028& 0.0140 \pm 0.0004& 0.0033 \pm 0.0001\\
\textbf{Scratch} & 97.04 \pm 0.15 & 98.00 & 0.0316\pm 0.0023 & 0.0278 \pm 0.0020 & 0.0917\pm 0.0126 & 0.0506 \pm 0.0238& 0.1506 \pm 0.0130& 0.0159 \pm 0.0009\\
\textbf{Scratch (Diff. Init.)} &  97.19 \pm 0.33 & 98.43 & 0.0352\pm 0.0037 & 0.0278 \pm 0.0032 & 0.1358\pm 0.0202 & 0.0650 \pm 0.0301& 0.1468 \pm 0.0295& 0.0157 \pm 0.0021\\
\textbf{Prune Restart } & 98.60 \pm 0.01 & 98.63 & 0.0027\pm 0.0003 & 0.0077 \pm 0.0003 & 0.0014\pm 0.0001 & 0.0019 \pm 0.0014& 0.0105 \pm 0.0001& 0.0027 \pm 0.0000\\
\midrule
\textbf{Pruned w/ Diff. Init.\tnote{**}}& 98.30 \pm 0.23 & 99.07 & 0.0214\pm 0.0023 & 0.0197 \pm 0.0019\tnote{***} & 0.0741\pm 0.0107 & 0.0438 \pm 0.0241& 0.0752 \pm 0.0133& 0.0103 \pm 0.0010\\
\end{tabular}
\begin{tabular}{@{}lS[table-format=2.3(4)]S[table-format=2.2]S[table-format=1.4(4)]S[table-format=1.4(4)]S[table-format=5.1(3)]S[table-format=5.1(3)]S[table-format=7.2(3)]S[table-format=7.2(3)]@{}}
\midrule
\textbf{Init.\ Method} & \textbf{Test Acc. (Top-1)} & \textbf{Ensemble} & \textbf{Pairwise Disagree.} & \textbf{Disagree. w/ Pruned} & \textbf{Pairwise \acrshort{kl}} & \textbf{\acrshort{kl} w/ Pruned} & \textbf{5-model \acrshort{jsd}} & \textbf{\acrshort{jsd} w/ Pruned}\\
\midrule
\multicolumn{3}{l}{\imgnet{}/ResNet-50}&&&&&&\\
\midrule
\textbf{Pruned Soln.}\tnote{*} & 0.7554 & {--} & {--} & {--} & {--} & {--} & {--} & {--}\\
\textbf{\gls{lt}} & 75.73\pm0.0008 & 76.27 & 0.0894\pm0.0008 & 0.0941\pm0.0009 & 3325\pm26.4 & 3605\pm12.4 & 787.2\pm5.24 & 853.7\pm3.55\\
\textbf{Scratch} & 71.16\pm0.13 & 74.05 & 0.2039\pm0.0013 & 0.2033\pm0.0012 & 15787\pm105 & 20442\pm178 & 3316\pm13.2 & 3847\pm26.0\\
\midrule
\textbf{Pruned w/ Diff. Init.\tnote{**}}& 75.65\pm0.13 & 77.78 & 0.1620\pm0.0008 & 0.1623\pm0.0011\tnote{***} & 13158\pm143 & 13087\pm55.9 &2760\pm16.8 & 2755\pm4.96\\
\bottomrule
\end{tabular}
\begin{tablenotes}
\footnotesize
\item[*] This is the pruning solution that the \gls{lt} and scratch models are derived from.
\item[**] 5 pruning solutions found with different random initialization, one of which is the pruning solution above.
\item[***] Here we compare 4 different pruned models with the pruning solution the LT/Scratch are derived from.
\end{tablenotes}
\end{threeparttable}
\end{sidewaystable*}
\Cref{disagreementfull} gives a full list of comparison metrics of the predictions on the test set for LeNet5 on MNIST and ResNet50 on \imgnet{}, in particular here we also compare the output probability distributions using relevant metrics.

\Citet{Fort2020deep} motivate deep ensembles by empirically showing that models starting from different random initializations typically learn different solutions, as compared to models trained from similar initializations.
Here we adopt the analysis of \citep{Fort2020deep}, but in comparing \gls{lt} initializations and random initializations using \emph{fractional disagreement} --- the fraction of class predictions over which the LT and scratch models disagree with the pruning solution they were derived from. In \cref{disagreementfull} we show the mean fractional disagreement over all pairs of models. We run two versions of scratch training: (1) \textit{Scratch (Diff. Init.} different weight initialization and different data order (2) \textit{Scratch} same weight initialization and different data order for 5 different seeds the experiments are ran. Finally, we restart training starting from the pruning solution (\textit{Prune Restart}) using, again, 5 different data orders.
 
The results presented in \cref{disagreementfull} suggest that all 5 \glspl{lt} models converge on a solution almost identical to the pruning solution. Interestingly, the 5 \gls{lt} models are even more similar to each other (\textit{Disagree.} column) than the pruning solution, possibly because they share an initialization and training is stable~\citep{Frankle2019stabilizing}. The disagreement of \textit{Prune Restart} solutions with the original pruning solution matches the disagreement of lottery solutions; showing the extent of similarity between \gls{lt} and pruning solutions.

Our results show that having a fixed initialization alone can not explain the low disagreement observed for \gls{lt} experiments as \textit{Scratch} solutions obtain an average disagreement of $0.0316$ despite using the same initialization, which is almost 10 times more than the \gls{lt} solutions ($0.0043$). Finding different \gls{lt} initialization is costly, however using a different initialization in \textit{Scratch (Diff. Init.)} training is free as the initializations are random. Using different initializations we can obtain more diverse solutions and thus achieve higher ensemble accuracy.
As suggested by the analysis of \citet{Fort2020deep}, ensembles of different solutions are more robust, and generalize better, than ensembles of similar solutions. An ensemble of 5 \gls{lt} models with low disagreement doesn't significantly improve generalization as compared to an ensemble of 5 different pruning solutions with similar individual test accuracy. We further demonstrate these results by comparing the output probability distributions using the \gls{kl}, and \gls{jsd}.

\end{document}